\documentclass[10pt,twocolumn,letterpaper]{article}

\usepackage[pagenumbers]{wacv} 

\usepackage{amsmath}
\usepackage{bm}

\definecolor{wacvblue}{rgb}{0.21,0.49,0.74}
\usepackage[pagebackref,breaklinks,colorlinks,allcolors=wacvblue]{hyperref}

\def\wacvPaperID{2682} 
\def\confName{WACV}
\def\confYear{2026}

\title{Blur2Sharp: Human Novel Pose and View Synthesis with Generative Prior Refinement}

\author{
Chia-Hern Lai \qquad
I-Hsuan Lo \qquad
Yen-Ku Yeh \quad
Thanh-Nguyen Truong \qquad
Ching-Chun Huang\thanks{Corresponding author, \\Project page: \url{https://essen900718.github.io/Blur2Sharp}} \\
National Yang Ming Chiao Tung University \\
{\tt\small lchxx98.ee11@nycu.edu.tw, chingchun@nycu.edu.tw} 
}

\begin{document}
\maketitle

\newcommand{\acknowledgement}[1]{{
  \let\thempfn\relax 
  \footnotetext[0]{#1}
}}



\begin{abstract}
The creation of lifelike human avatars capable of realistic pose variation and viewpoint flexibility remains a fundamental challenge in computer vision and graphics. Current approaches typically yield either geometrically inconsistent multi-view images or sacrifice photorealism, resulting in blurry outputs under diverse viewing angles and complex motions. To address these issues, we propose Blur2Sharp, a novel framework integrating 3D-aware neural rendering and diffusion models to generate sharp, geometrically consistent novel-view images from only a single reference view. Our method employs a dual-conditioning architecture: initially, a Human NeRF model generates geometrically coherent multi-view renderings for target poses, explicitly encoding 3D structural guidance. Subsequently, a diffusion model conditioned on these renderings refines the generated images, preserving fine-grained details and structural fidelity. We further enhance visual quality through hierarchical feature fusion, incorporating texture, normal, and semantic priors extracted from parametric SMPL models to simultaneously improve global coherence and local detail accuracy. Extensive experiments demonstrate that Blur2Sharp consistently surpasses state-of-the-art techniques in both novel pose and view generation tasks, particularly excelling under challenging scenarios involving loose clothing and occlusions.

\end{abstract}
    
\section{Introduction}
\label{sec:intro}

Generating realistic human avatars capable of novel views and diverse poses remains a fundamental challenge in computer vision, with extensive implications for next-generation technologies such as virtual reality (VR), augmented reality (AR), gaming, and commercial applications like virtual try-on. High-fidelity digital humans that exhibit natural motion, consistent geometric structure, and seamless integration into dynamic environments are essential for immersive content creation and interactive experiences. Nevertheless, synthesizing photorealistic human avatars while ensuring geometric consistency across varied viewpoints and complex poses is still an open research problem, significantly influencing the development of interactive media and virtual fashion industries.

\begin{figure*}[t]
  
  \includegraphics[scale=0.33]{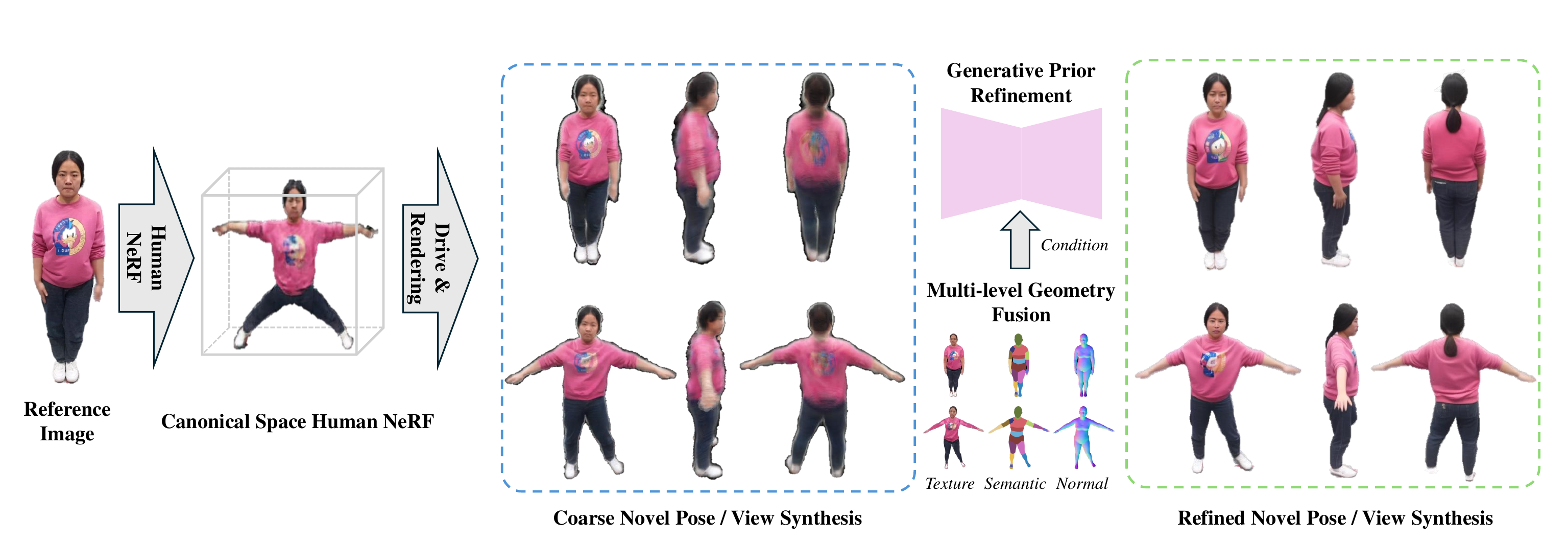}  
  \centering
  \caption{Based on a single reference image, the proposed \textbf{Blur2Sharp} significantly enhances photorealism in novel pose and view synthesis. It introduces a generative refinement module conditioned on multi-level geometric information to effectively resolve the coarse and blurred artifacts typically observed in previous approaches~\cite{sherf}, thus achieving sharper visual quality and robust geometric consistency.}
  \label{fig: Overview}  
\end{figure*}

Recent advances in Neural Radiance Fields (NeRF) \cite{nerf} have substantially improved the ability to synthesize 3D human avatars from sparse image collections. To enhance novel-view generalization across varying appearances and postures, related approaches \cite{sherf} typically employ a learned canonical representation to align diverse subjects and poses within a common latent space. Despite these successes, these regression-based methods exhibit notable limitations. Their deterministic formulations often result in oversmoothed outputs, particularly when handling complex deformations such as loose clothing or hair and synthesizing novel-view appearances.

On the other hand, the advances in video diffusion models \cite{video_diffusion,guo2023animatediff}, driven by large-scale training datasets, have significantly improved the video generation of human animations from just a single reference image. Several methods \cite{dispose,AnimateAnyone,magicanimate,unianimate,magicpose} leverage generative priors to synthesize realistic animations, offering fine-grained control guided by 2D keypoints \cite{dw-pose,openpose}. To enhance motion control further, Champ \cite{Champ} integrates the 3D parametric human model SMPL \cite{SMPL}, incorporating additional pose information via normal and semantic maps. However, video diffusion models inherently lack explicit 3D reasoning capabilities. This limitation results in multi-view inconsistencies, including texture flickering and geometric misalignments when observed from novel viewpoints.

In the field of single-image generative novel-view synthesis, recent approaches such as CharacterGen \cite{charactergen} and MagicMan \cite{magicman} have aimed to improve multi-view consistency by incorporating viewpoint control signals and multi-view feature aggregation. While these methods successfully enhance standard diffusion-based synthesis by explicitly enforcing cross-view correspondences, they remain limited in scope. For instance, CharacterGen enables character animation through skeleton-based rigging, but its core pipeline primarily focuses on reconstructing static human appearances before applying motion. Consequently, these methods lack the capability to directly synthesize dynamic content with novel poses, limiting their applicability to scenarios that demand continuous pose variation, such as realistic virtual avatars.

To address these challenges, we propose \textbf{Blur2Sharp}, a novel generative framework for synthesizing lifelike human avatars from a single reference image, capable of generating both novel views and novel poses. Blur2Sharp bridges the gap between 3D-aware reconstruction and diffusion-based synthesis, leveraging their complementary strengths. As illustrated in Fig.~\ref{fig: Overview}, our approach employs a hybrid pipeline that integrates a NeRF-based human reconstruction model and a diffusion-based generative refinement module. Specifically, we first generate geometrically consistent multi-view renderings under the desired target poses using a NeRF-based 3D human model. These coarse outputs are subsequently refined using a diffusion model guided by parametric human priors, significantly enhancing realism and achieving photorealistic quality.

The key innovation of our framework lies in its novel \textbf{conditioning mechanism}, where initial NeRF renderings provide complementary guidance throughout the diffusion process. This design enables geometry-aware generation while preserving detailed textures and visual realism. 
Additionally, we introduce a \textbf{multi-layer feature fusion strategy} that progressively integrates texture, normal, and semantic priors at multiple network stages. This 
integration facilitates precise control over global structural coherence and detailed surface features simultaneously. By effectively leveraging these multi-level conditioning signals, \textbf{Blur2Sharp} demonstrates robust generalization to diverse subjects and poses while maintaining spatial coherence and pose consistency.

In summary, the key contributions of this work include:
\begin{enumerate} 
\item We propose \textbf{Blur2Sharp}, a novel generative framework that integrates novel-view synthesis and pose manipulation, enabling the creation of high-quality multi-view, multi-pose human images from only a single reference image.
\item We introduce a generative prior refinement module, which conditions the diffusion process on NeRF renderings, effectively balancing geometric fidelity and photorealistic detail.
\item We develop a novel guidance strategy that fuses texture, normal, and semantic priors for fine-grained control over geometry and appearance throughout the diffusion process.
\end{enumerate}

\section{Related Work}
\label{sec:formatting}


\subsection{3D Human Avatar Modeling with few inputs}

Traditional human synthesis approaches using Neural Radiance Fields (NeRF) have predominantly relied on multi-view image collections or monocular video sequences captured from fixed viewpoints. Early methods~\cite{su2021anerf, neuralbody} modeled 3D humans using deformation fields derived from skeletal pose estimations or sparse 3D convolutions, but these approaches require extensive per-subject optimization, significantly restricting their generalization capabilities. To overcome these limitations, recent methods~\cite{mu2023actorsnerf,mpsnerf,weng2022humannerf, peng2021animatable, te2022neuralcapture} developed pose-dependent generalizable NeRFs to map diverse human subjects into a shared canonical representation and enabling efficient single-pass inference. Among these, SHERF~\cite{sherf} achieves impressive generalization from single-view inputs through pixel-aligned and point-wise feature fusion.

More recently, 3D Gaussian Splatting~\cite{gaussian_splatting} has emerged as a promising, efficient alternative to traditional NeRF-based rendering, enabling rapid training and real-time visualization by representing scenes with compact 3D Gaussians. Several recent techniques~\cite{kwon2024generalizablehumangaussianssparse,humansplat,gst,idol} have extended this method to support novel poses and viewpoints. However, all these methods often struggle to produce sharp and detailed results due to their modeling nature. In our work, we bridge this gap by combining the robust geometric understanding from Human NeRF with generative priors, achieving multi-view consistency and sharp photorealistic detail in human avatar synthesis.

\subsection{Generative Human Image Animation}

Human image animation has seen significant progress through generative models, enabling motion synthesis from a single reference image. Recently diffusion models~\cite{unianimate,mimicmotion2024,dreampose} provide superior image quality through iterative denoising. Animate Anyone~\cite{AnimateAnyone} introduces a reference UNet to preserve subject identity throughout the animation process, complemented by a pose extractor module for integrating skeletal information. Building on this, several methods~\cite{magicanimate,dispose,magicpose} adopt similar strategies for novel pose synthesis. Champ~\cite{Champ} incorporates a parametric 3D human body model to impose additional structural constraints, while DisPose~\cite{dispose} enhances temporal consistency using motion field guidance with keypoint correspondences. Despite these advancements, existing methods still struggle with multi-view consistency due to limited 3D awareness~\cite{anigs}. To address this, we integrate NeRF-based appearance and normal priors into diffusion models, enabling geometrically consistent and photorealistic avatar generation across diverse poses and viewpoints.

\subsection{Generative Novel View Synthesis}

Diffusion models have recently achieved remarkable progress in novel view synthesis, enabling the generation of coherent images from diverse viewpoints. Early approaches, such as Zero-1-to-3~\cite{liu2023zero1to3}, fine-tune 2D diffusion models for view synthesis by conditioning on relative camera poses. However, these methods rely solely on 2D supervision, making it difficult to preserve full 3D consistency. To improve 3D coherence, methods like Score Distillation Sampling (SDS)\cite{poole2022dreamfusion} optimize 3D representations using diffusion priors, but often suffer from artifacts 
due to noisy gradients. For human-specific synthesis, CharacterGen\cite{charactergen} introduces cross-view attention mechanisms to align features across different viewpoints, while MagicMan~\cite{magicman} further advances this direction by jointly generating aligned color images and normal maps through a unified diffusion framework. Despite these advances, most of these methods focus on static objects or human subjects from novel viewpoints, without explicitly addressing pose variations. In contrast, our work builds on these foundations by introducing a multi-view diffusion framework capable of synthesizing both novel viewpoints and novel poses of humans, with joint generation in both RGB and normal domains.

\section{Method}

\begin{figure*}[t]
  
  \includegraphics[scale=0.5]{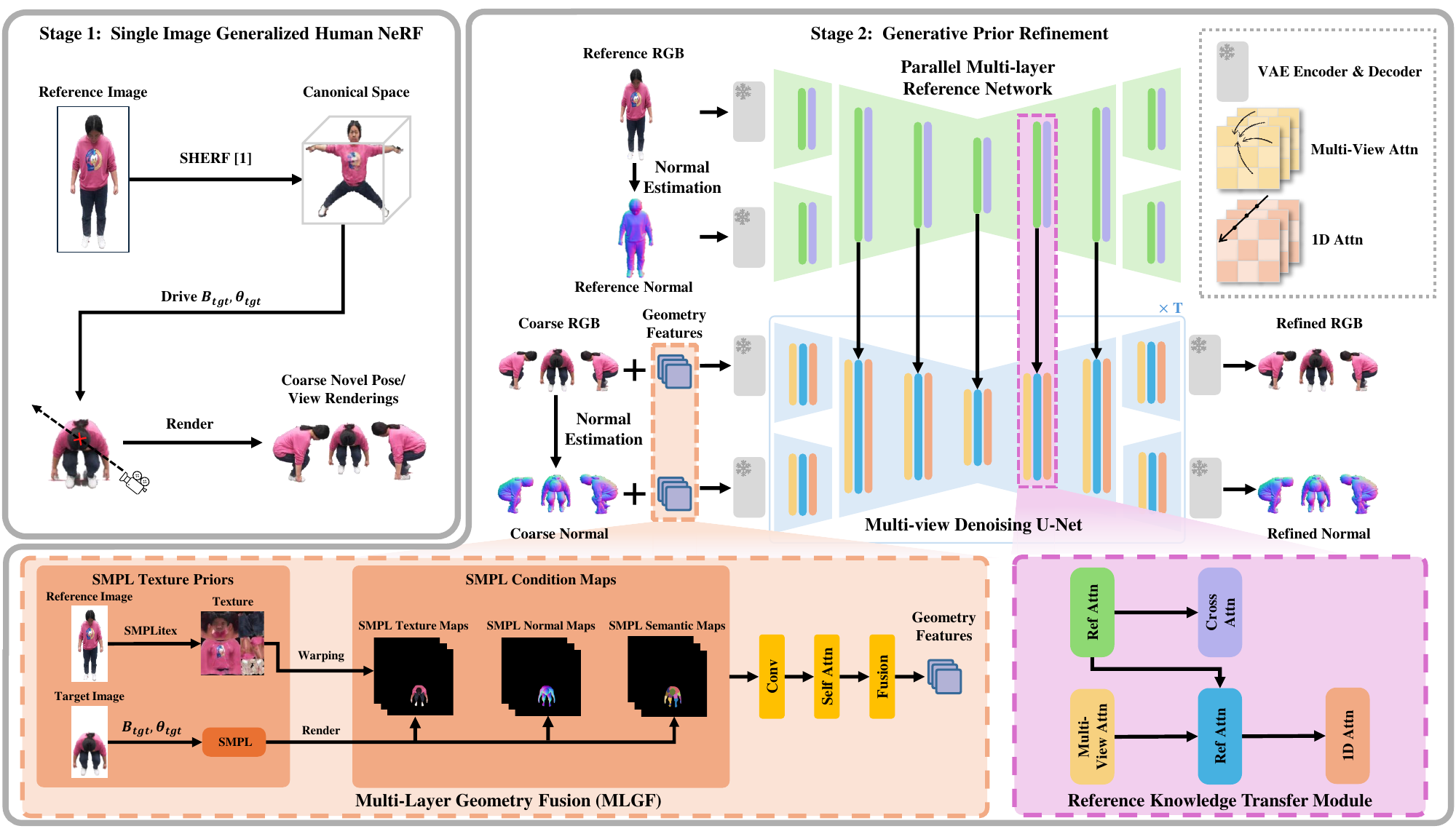} 
  \centering
  \caption{\textbf{System architecture.} Our framework operates in two main stages: (1) a generalizable Human NeRF module~\cite{sherf} that maps the reference image to a canonical space to generate initial novel view renderings under target body parameters \(\bm{\beta}_{\text{tgt}}\) and pose parameters \(\bm{\theta}_{\text{tgt}}\); and (2) a refinement network that leverages a multi-view denoising U-Net conditioned on the NeRF's RGB renderings, estimated normal maps, and additional geometric features to produce enhanced RGB and Normal maps. Specifically, the MLGF module generates the geometric features by fusing the SMPL texture priors with two SMPL geometric priors (normal and semantic maps). To further preserve subject-specific details, the refinement network incorporates a Reference Knowledge Transfer Module that propagates features from the reference image and normal map via reference attention.}
  \label{fig:archictecture_2} 
\end{figure*}

\subsection{Method Overview}
We present \textbf{Blur2Sharp}, a novel framework for human pose and view synthesis from a single reference image, which combines a generalized human NeRF with a generative refinement module. Our goal is to synthesize realistic human images that faithfully preserve the subject’s appearance while accurately conforming to specified target poses and camera viewpoints.

Given a reference image \( I_{\text{ref}} \), along with its associated camera parameters \( \mathbf{c}_{\text{ref}} \), pose parameters \( \bm{\theta}_{\text{ref}} \in \mathbb{R}^{72} \), and body shape parameters \( \bm{\beta}_{\text{ref}} \in \mathbb{R}^{10} \), our framework generates images of the subject in a target pose \( \bm{\theta}_{\text{tgt}} \), target body shape \( \bm{\beta}_{\text{tgt}} \), and from a novel camera viewpoint \( \mathbf{c}_{\text{tgt}} \). The pipeline operates in two stages as illustrated in Fig.~\ref{fig:archictecture_2}. In the first stage, a Human NeRF produces coarse renderings conditioned on the target pose, shape, and camera parameters, maintaining visual consistency with the reference image (see Sec.~\ref{sec:Single-Image Generalizable Human NeRF}). In the second stage, these coarse outputs are refined using a generative model that enforces multi-view and pose consistency (see Sec.~\ref{sec:Generative Prior Refinement}) and, guided by texture and geometric priors derived from the SMPL body shape representation~\cite{SMPL,smplitex} (see Sec.~\ref{sec:Multi-Layer Geometry Guidance}).

During training, we first train a Human NeRF model that can represent the target subject in 3D and generate coarse renderings for given camera views, novel target poses, and the reference image. We then train a diffusion-based generative model consisting of a multi-view denoising UNet and a reference-guided UNet to refine these renderings. Full training details are provided in Supplementary Materials Sec. A.2 and A.3.

\subsection{Human NeRF Construction Using a Single Reference Image}
\label{sec:Single-Image Generalizable Human NeRF}
Here, we follow SHERF~\cite{sherf} to train a generalized Human NeRF model, \( F_\phi \), capable of synthesizing novel views of a human subject from a single reference image under arbitrary poses. Given a reference image \( I_{\text{ref}} \), the trained model \( F_\phi \) renders the target image of the subject from a novel camera viewpoint \( c_{\text{tgt}} \), conditioned on the target SMPL body shape \( \bm{\beta}_{\text{tgt}} \) and pose parameters \( \bm{\theta}_{\text{tgt}} \). 

Specifically, the rendering process begins by casting rays from the target camera view \( c_{\text{tgt}} \) through each pixel in the output image. Along each ray, a set of 3D points \( \mathbf{x}_{\text{tgt}} \) is sampled in the target SMPL space. To transfer appearance information from the reference image \( I_{\text{ref}} \) to the target SMPL model, these 3D target points are mapped to the canonical SMPL space using inverse LBS, resulting in the corresponding canonical SMPL points \( \mathbf{x}_{\text{c}} \). The features \( \rho_{\text{tgt} \leftarrow \text{ref}} \) of these target points are then retrieved by mapping the \textbf{reference SMPL} points to the \textbf{canonical SMPL} space.

The feature extraction of \( \rho_{\text{tgt} \leftarrow \text{ref}} \) follows a two-step approach. First, we compute pixel-aligned features by projecting each canonical point $\mathbf{x}_c$ to the reference SMPL space via forward LBS to obtain \( \mathbf{x}_{\text{ref}} \), which are then projected onto the reference image feature map to extract fine-grained appearance details from the reference image \( I_{\text{ref}} \). Second, the 3D point-level features, \( \rho_{\text{tgt} \leftarrow \text{ref}} \) are extracted by sparse 3D convolutions of the pixel-aligned features to recover 3D human structure and local appearance details. Then, the extracted point features is passed through the trained NeRF MLP network $F_\phi$ to predict the density $\sigma(\mathbf{x}_{\text{tgt}})$ and color $c(\mathbf{x}_{\text{tgt}})$ at each sampled target point by
\begin{equation}
    \sigma(\mathbf{x}_{\text{tgt}}), \mathbf{c}(\mathbf{x}_{\text{tgt}}) = F_\phi\left(\mathbf{x}_{\text{tgt}}, \mathbf{d}, \rho_{\text{tgt} \leftarrow \text{ref}}\right),
\end{equation}
where $\mathbf{d}$ and $\rho_{\text{tgt} \leftarrow \text{ref}}$ denote the viewing direction to the target point $\mathbf{x}_{\text{tgt}}$ and the point feature respectively. The predicted densities and colors are aggregated through volume rendering to generate the final image in the desired target pose and camera view.

\subsection{Generative Prior Refinement}
\label{sec:Generative Prior Refinement}
While the single-image Human NeRF demonstrates strong capabilities in reconstructing the geometry of a target pose, it still faces significant challenges in generating high-fidelity appearance, particularly in the presence of occlusions. The core issue arises from the inherently underconstrained nature, where multiple plausible solutions exist for the occluded regions. As a result, the Human NeRF model often resorts to averaging pixel values across these ambiguities, leading to outputs that lack sharpness and fine detail. To address these limitations, we introduce a refinement stage that leverages diffusion-based generative priors to enhance the coarse renderings produced by the Human NeRF.

\paragraph{Coarse Image Conditioning.}To improve photorealism, we utilize the rendered coarse RGB outputs and the corresponding normal maps predicted by the off-the-shelf normal estimator~\cite{khirodkar2024sapiens} as global appearance and surface geometry priors for our refinement diffusion model. Given the coarse RGB images \(\{ I_{\text{coarse}}^i \}_{i=1}^{n} \) and the corresponding normal maps \(\{ N_{\text{coarse}}^i \}_{i=1}^{n} \), where each \( I_{\text{coarse}}^i \in \mathbb{R}^{H \times W \times 3} \) and \( N_{\text{coarse}}^i \in \mathbb{R}^{H \times W \times 3} \) represent the $i$th viewpoint, we encode them using a pretrained VAE encoder to obtain compressed conditional latent representations. Next, we use lightweight convolutional encoders to process these latent representations into condition RGB features \( \mathbf{F}_{\text{rgb}}^i \) and normal features \( \mathbf{F}_{\text{normal}}^i \). We then inject the noisy \( \mathbf{F}_{\text{rgb}}^i \) and \( \mathbf{F}_{\text{normal}}^i \) into the trained diffusion denoising network independently by combining them with the corresponding latent noises via elemental-wise addition for detail enhancement. For additional details regarding our conditioning mechanism, please refer to Supplementary Materials Sec. A.4.


\paragraph{Multi-view Denoising U-Net.} Inspired by recent advances in diffusion-based multi-view human synthesis~\cite{magicman, charactergen}, we adopt a multi-view denoising U-Net to enhance cross-view feature aggregation by integrating both 1D and multi-view attention mechanisms. The 1D attention operates along the view dimension, comparing features only at identical pixel locations across different views. While it is memory-efficient, it cannot account for spatial misalignments caused by viewpoint changes. To address this, we apply multi-view attention to combine features across different views, enabling the model to align and integrate information from corresponding regions that may be spatially shifted due to viewpoint changes. As the 1D attention already establishes initial inter-view connections, multi-view attention is applied selectively to a sparse set of views, effectively balancing memory efficiency and cross-view feature integration.

\paragraph{Reference Knowledge Transfer Module.} Similar to Animate Anyone~\cite{AnimateAnyone}, we adopt a parallel multi-layer reference network that mirrors the architecture of the multi-view denoising UNet, excluding the multi-view and 1D attention layers. This reference network encodes features from the input image and transfers them to the denoising network as reference knowledge, thereby preserving the distinctive visual characteristics of the reference subject throughout the generation process. To enable reference knowledge transfer, we replace the standard cross-attention layers in the denoising network and the self-attention layers in the reference network with reference attention~\cite{magicman}. Please refer to Supplementary Materials Sec. A.4 for further details on the Reference Knowledge Transfer Module.

\paragraph{Dual-Branch RGB-Normal Conditional Diffusion Model.}We extend a unified diffusion architecture inspired by \cite{magicman} to jointly model photorealistic appearance and 3D surface geometry, ensuring accurate spatial alignment between synthesized images and their normal maps. To be specific, we modify a standard UNet for Dual-Branch RGB-Normal denoising by duplicating the initial downsample and final upsample layers. Please refer to Supplementary Materials Sec. A.4 for further details.

To incorporate normal-based conditioning, we leverage an off-the-shelf monocular normal estimator~\cite{khirodkar2024sapiens} to derive two key inputs for our diffusion framework: (1) A reference normal map derived directly from the input RGB image, and (2) predicted normal maps derived from the coarse RGB outputs generated by the Human NeRF model. For diffusion refinement, We extract normal features, denoted as \( \mathbf{F}_{\text{normal}} \), using the same lightweight 2D convolutional architecture employed for RGB features \( \mathbf{F}_{\text{rgb}} \), but with separate parameters to allow independent learning. 

\subsection{Multi-Layer Geometry Guidance}
\label{sec:Multi-Layer Geometry Guidance}
\paragraph{SMPL Texture Priors.}Novel view and pose synthesis of humans from a single image presents a significant challenge, particularly in recovering appearance information from unseen perspectives. Prior works primarily rely on structural conditioning signals (e.g., SMPL normal maps or skeleton maps) to guide the denoising process during human generation. However, these signals lack the subject-specific appearance information necessary for high-fidelity synthesis. Coarse image renderings from Human NeRF are also insufficient to capture accurate appearance in novel views due to their inherent blurriness. Furthermore, the reference image provides only partial observations of the subject, omitting critical texture details from unobserved viewpoints. To address these limitations, we introduce SMPL texture priors to complement the missing texture information in these inputs.

To accomplish this, we begin by leveraging a SMPL representation derived from an existing pose estimation framework~\cite{SMPL}. Given a reference image, we construct a partial texture map \( UV_{\text{part}} \) by projecting the reference image onto the SMPL model's UV space. This process follows the standard technique of SMPLitex~\cite{smplitex}, a widely used method for capturing partial textures from a single image by establishing accurate pixel-to-surface correspondences. Then, we compute a human silhouette to remove background pixels and focus specifically on the body areas. The pixel-to-surface correspondence map \( d \) is masked by the silhouette \( s \), and the partial texture map is projected as:
\begin{equation}
UV_{\text{part}} = \Pi(I_{\text{ref}}, d \odot s),
\end{equation}
where \( \odot \) denotes the Hadamard product and \( \Pi \) projects image pixels to UV coordinates on the SMPL model. To handle missing or incomplete texture regions, we apply a diffusion-based inpainting model~\cite{latent_diffusion} to complete the partial texture map, filling in the gaps and generating a more complete texture representation for the SMPL model. The inpainted texture \( UV_{\text{inpaint}} \) is then wrapped onto the 3D surface of the target SMPL model. The texture wrapping process is expressed as:

\begin{equation}
C_{\text{texture}}^i = \mathcal{W}(UV_{\text{inpaint}}, \beta_{\text{tgt}}, \theta_{\text{tgt}}, c_i),
\label{equation_3}
\end{equation}

where \( \mathcal{W} \) is the warping function that maps the inpainted texture onto the 3D SMPL surface, followed by rendering with camera parameters \( c_i \) to obtain the final projected texture \(C_{\text{texture}}^i \).

\paragraph{Multi-Layer Geometry Fusion (MLGF).} To further enhance the final novel pose and view generation, we integrate three SMPL condition maps—texture, normal, and semantic prior maps to construct the geometry features used in the generative refinement process. The texture map is obtained using Equation~\ref{equation_3}, while the normal and semantic maps are obtained by projecting surface normals and predefined part labels of the target SMPL mesh onto the image plane. For each viewpoint \( i \), the texture, normal, and semantic condition maps are independently processed through a lightweight network composed of four convolutional layers, followed by a self-attention layer with zero convolution. The resulting geometry features are then fused by summing the outputs from each modality to produce the geometry feature \( \mathbf{F}_{\text{geo}}^i \) for viewpoint \( i \):

\begin{equation}
\mathbf{F}_{\text{geo}}^i = \sum_{k} \text{SelfAttn}_k(\text{ConvBlock}_k(\mathbf{C}_k^i)),
\end{equation}

where $\mathbf{C}_k^i$ denotes the condition maps and \( k \) indicates the
condition types (i.e., texture, normal, and semantic). The fused geometry features are subsequently combined with latent noise representations and integrated with both RGB and normal latents via element-wise addition, which are subsequently used for generating the refined images and normal maps.

\section{Experiments}

\subsection{Dataset}
We evaluate on two multi-view video datasets: HuMMan and MVHumanNet dataset. For HuMMan dataset, we use 1{,}695 sequences from 339 subjects, each with 5 target frames from 4 views and 1 front-view reference. We split 1{,}530 for training and 165 for testing. For MVHumanNet, we sample 4{,}715 sequences from 943 subjects using the same setup, with 4{,}432 for training and 283 for testing.
Both datasets provide annotated foreground masks, SMPL parameters, and calibrated camera parameters to facilitate evaluation. Also, we generate normal maps for all images using an off-the-shelf method~\cite{khirodkar2024sapiens} to support our analysis.

\subsection{Comparison Methods and Evaluation Metrics} We evaluate against state-of-the-art single-image human generation methods, including the generalizable NeRF approach SHERF~\cite{sherf} and diffusion-based methods Animate Anyone~\cite{AnimateAnyone} and Champ~\cite{Champ}. For a fair comparison, SHERF and Champ are re-trained using their official implementations, while Animate Anyone is reproduced via open-source codebase from MooreThreads under identical training data and settings.
Our evaluation covers two core tasks: novel pose synthesis across multiple views and novel view synthesis. For both tasks, we quantify performance using PSNR, SSIM, LPIPS, and FID, with ground-truth multi-view images.

\subsection{Novel Pose Synthesis across Multiple Views}

\begin{figure*}
    \includegraphics[scale=0.51]{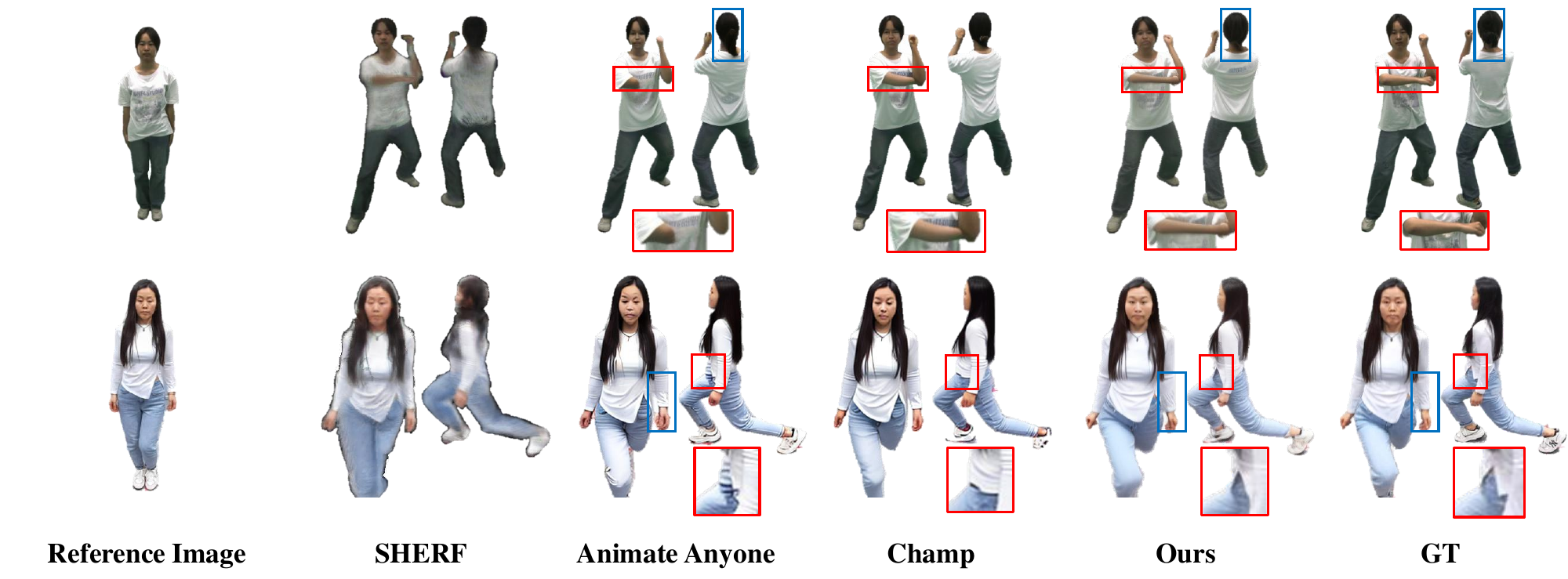}  
  \centering
    \caption{Qualitative comparisons of novel pose synthesis from multiple views on MVHumanNet and HuMMan, showcasing our method alongside SHERF, Animate Anyone, and Champ. Overall, our method yields more accurate poses and more consistent appearance than prior methods. \textcolor{red}{Red} boxes indicate enlarged regions, while \textcolor{blue}{blue} boxes highlight areas with additional artifacts.}
  \label{fig: novel_pose_figure} 
\end{figure*}

\begin{table}[]
\centering
\caption{Quantitative comparison of novel pose synthesis across 4 views on the MVHumanNet and HuMMan datasets. The evaluation metrics include PSNR, SSIM, LPIPS, and FID. The top two methods for each dataset are highlighted in \textbf{bold} and \underline{underline}.}
\resizebox{\columnwidth}{!}{%
\begin{tabular}{l|cccc|cccc}
\toprule
Method & \multicolumn{4}{c|}{MVHumanNet} & \multicolumn{4}{c}{HuMMan} \\
 & PSNR $\uparrow$ & SSIM $\uparrow$ & LPIPS $\downarrow$ & FID $\downarrow$ & PSNR $\uparrow$ & SSIM $\uparrow$ & LPIPS $\downarrow$ & FID $\downarrow$ \\
\midrule
SHERF & 20.27 & 0.929 & 0.059 & 91.42 & 21.22 & 0.909 & 0.078 & 90.85 \\
\midrule
Animate Anyone & 22.19 & 0.941 & 0.043 & 28.95 & 24.80 & 0.934 & 0.042 & 32.12 \\
Champ & \underline{22.25} & \underline{0.941} & \underline{0.042} & \underline{27.09} & \underline{25.12} & \underline{0.937} & \underline{0.041} & \underline{30.91} \\
\midrule
\textbf{Ours (Blur2Sharp)} & \textbf{23.31} & \textbf{0.946} & \textbf{0.039} & \textbf{24.38} & \textbf{26.60} & \textbf{0.946} & \textbf{0.034} & \textbf{22.31} \\
\bottomrule
\end{tabular}%
}
\label{tab:novel_pose_results}
\end{table}

In this task, we focus on generating novel pose synthesis in a multi-view setting rather than within the same viewpoint. This approach better aligns with real-world scenarios, where the appearance of a person varies across different perspectives. We compare our method, Blur2Sharp, with three state-of-the-art generative human synthesis methods: Animate Anyone~\cite{AnimateAnyone}, Champ~\cite{Champ}, and the single-image generalized human NeRF model, SHERF~\cite{sherf}. For a fair comparison, we train Animate Anyone and Champ for 30k iterations on each dataset, while SHERF is trained for 80k iterations.

Tab.~\ref{tab:novel_pose_results} shows that our method outperforms all baselines across all metrics. SHERF performs the worst results due to its blurry outputs, especially for unseen views and poses, as it struggles to infer missing details from a single image. Fig.~\ref{fig: novel_pose_figure} illustrates qualitative comparisons. While Animate Anyone and Champ produce reasonable perceptual quality in unseen views and poses, they struggle with fine-grained details and fail to maintain multi-view consistency. SHERF suffers from significant blurriness and often generates incorrect appearances, especially in back views. Our method effectively preserves fine details while ensuring multi-view consistency, demonstrating superior performance in novel pose synthesis across multiple perspectives.

\subsection{Novel View Synthesis}

\begin{figure*}[!t]
  \includegraphics[scale=0.5]{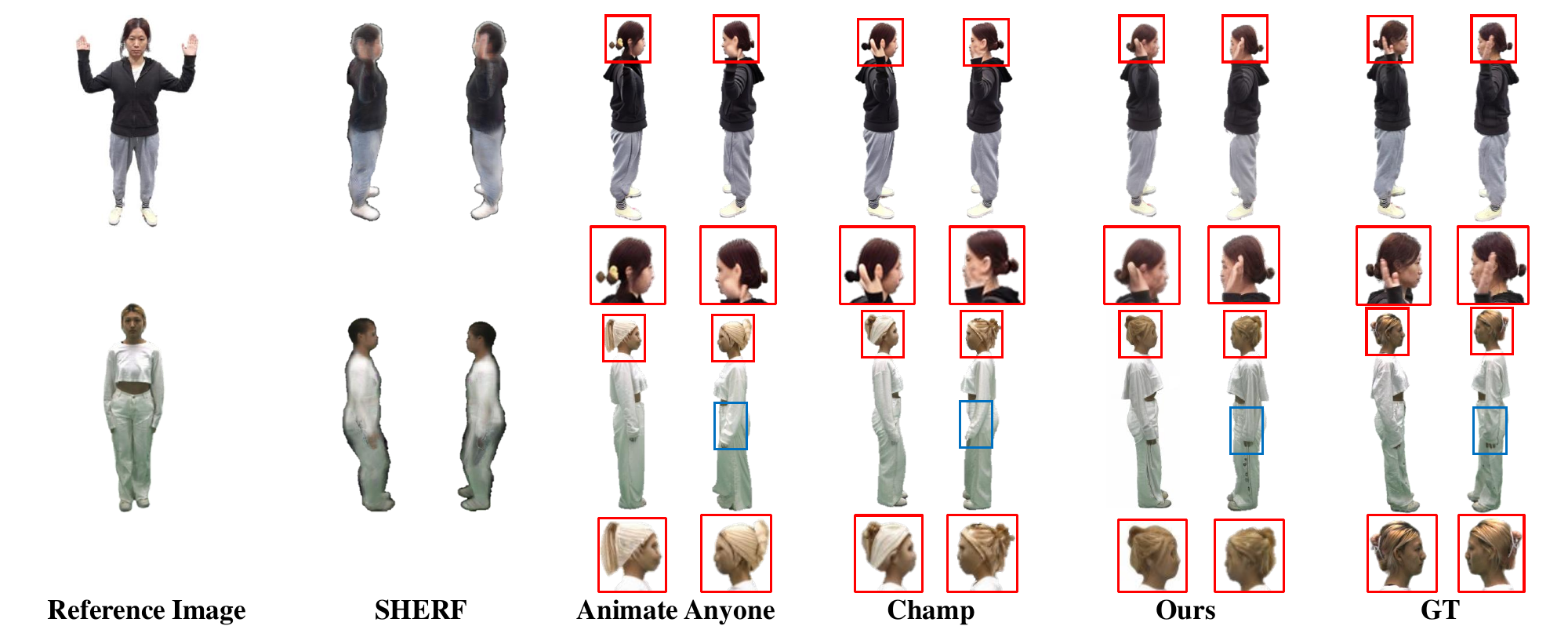}  
  \centering
    \caption{Qualitative results of novel view synthesis across 4 views on MVHumanNet and HuMMan dataset. Our method shows improved realism and consistency with fewer artifacts. \textcolor{red}{Red} boxes indicate enlarged regions, while \textcolor{blue}{blue} boxes highlight areas with additional artifacts.}
  \label{fig: novel_view_figure}  
\end{figure*}

\begin{table}[t]
\centering
\caption{Quantitative comparison of novel view synthesis across 4 views on the MVHumanNet and HuMMan datasets. The top two methods for each dataset are highlighted in \textbf{bold} and \underline{underline}.}
\resizebox{\columnwidth}{!}{%
\begin{tabular}{l|cccc|cccc}
\toprule
Method & \multicolumn{4}{c|}{MVHumanNet} & \multicolumn{4}{c}{HuMMan} \\

 & PSNR $ \uparrow$ & SSIM $ \uparrow$ & LPIPS $ \downarrow$ & FID $ \downarrow$ & PSNR $ \uparrow$ & SSIM $ \uparrow$ & LPIPS $ \downarrow$ & FID $ \downarrow$ \\
\midrule
SHERF & 20.81 & 0.935 & 0.054 & 83.34 & 21.79 & 0.917 & 0.073 & 88.73\\
\midrule
Animate Anyone & 22.64 & 0.943 & 0.040 & \underline{27.86} & 25.59 & 0.939 & 0.038 & 30.78 \\
Champ & \underline{22.98} & \underline{0.946} & \underline{0.038} & 29.36 & \underline{25.86} & \underline{0.942} & \underline{0.038} & \underline{29.67}  \\
\midrule
\textbf{Ours (Blur2Sharp)} & \textbf{23.82} & \textbf{0.948} & \textbf{0.036} & \textbf{23.50} & \textbf{27.01} & \textbf{0.948} & \textbf{0.033} & \textbf{22.21} \\
\bottomrule
\end{tabular}%
}
\label{tab:novel_view_results}
\vspace{-1em}
\end{table}

For the novel view synthesis task, all methods are evaluated under consistent conditions using the same reference frame. Blur2Sharp outperforms baseline approaches in both geometric consistency and perceptual sharpness. Tab.~\ref{tab:novel_view_results} shows that Blur2Sharp achieves superior quantitative results across all metrics. Fig.~\ref{fig: novel_view_figure} highlights that our method produces photorealistic and consistent novel views, while other methods struggle with blurriness, detail loss, or view inconsistencies.

\subsection{Ablation Studies}

\begin{table}[]
\centering
\caption{Quantitative ablation study on MVHumanNet dataset. 
}
\resizebox{\columnwidth}{!}{%
\begin{tabular}{l|cccc|cccc}
\toprule
Method & \multicolumn{4}{c|}{Novel Pose Synthesis} & \multicolumn{4}{c}{Novel View Synthesis} \\

 & PSNR $ \uparrow$ & SSIM $ \uparrow$ & LPIPS $ \downarrow$ & FID $ \downarrow$ & PSNR $ \uparrow$ & SSIM $ \uparrow$ & LPIPS $ \downarrow$ & FID $ \downarrow$ \\
\midrule
Ours  (w/o. normal) & 23.10 & 0.945 & 0.040 & 30.61 & 23.45 & 0.947 & 0.038 & 30.04 \\
Ours  (w/o. MLGF) & 22.94 & 0.943 & 0.041  & 27.59 & 23.51 & 0.946 & 0.038 & 26.69 \\
Ours  (w/o. Human NeRF) & \underline{23.20} & \underline{0.945} & \underline{0.039} & \underline{25.41} & \underline{23.60} & \underline{0.947} & \underline{0.037} & \underline{25.05}  \\
\textbf{Ours} & \textbf{23.31} & \textbf{0.946} & \textbf{0.039} & \textbf{24.38} & \textbf{23.82} & \textbf{0.948} & \textbf{0.036} & \textbf{23.50}  \\
\bottomrule
\end{tabular}
}
\label{tab:ablation_study}
\end{table}

\begin{figure}[]
  \includegraphics[scale=0.35]{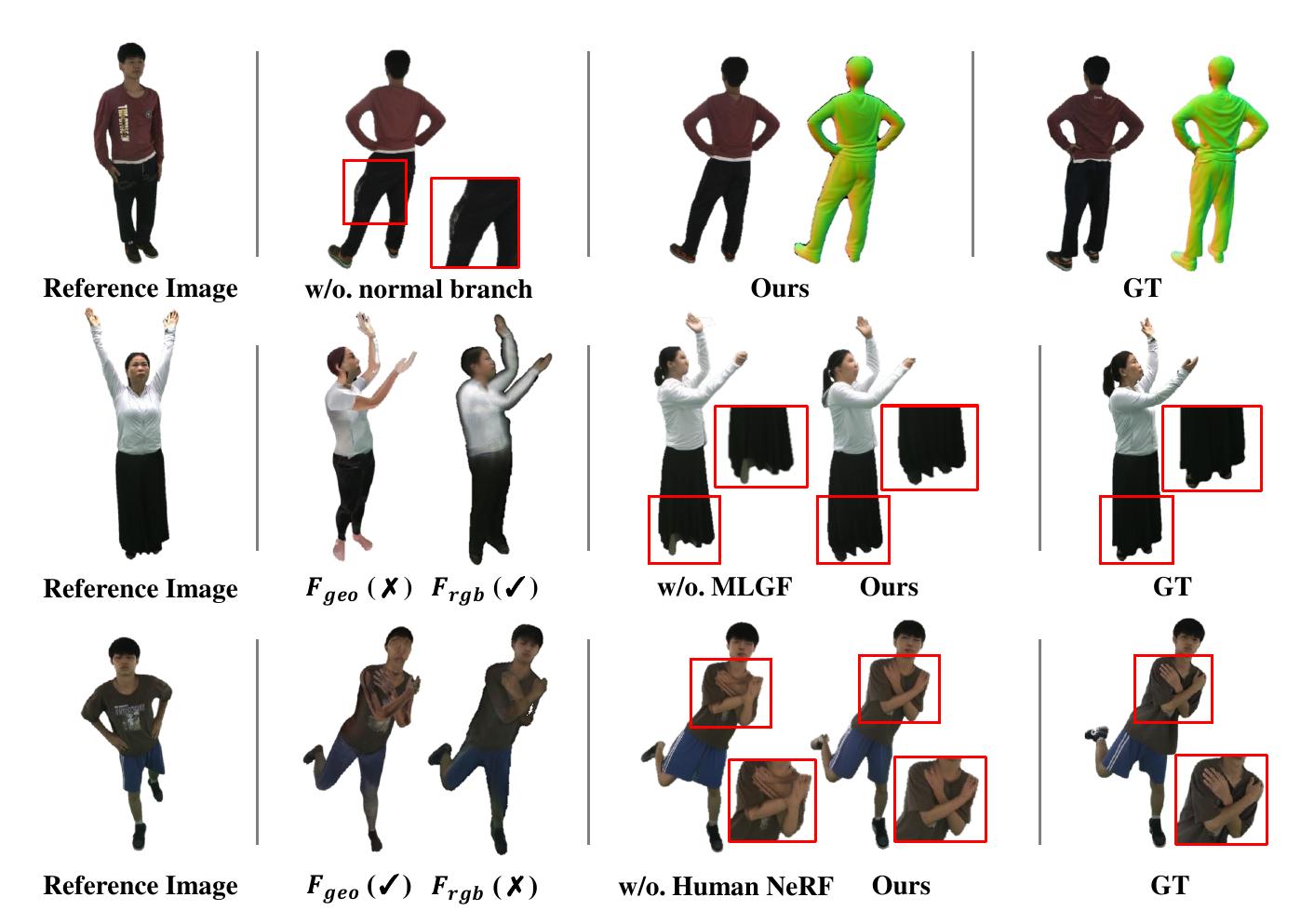}  
  \centering
    \caption{Qualitative ablation study on MVHumanNet dataset. $F_{geo}$ and $I_{coarse}$ denotes geometry features and image rendered by Human NeRF respectively. \textcolor{red}{Red} boxes indicate enlarged regions.}
  \label{fig:ablation}  
\end{figure}

\paragraph{Dual-domain Diffusion Model.}To validate the importance of our dual-domain approach, we conduct an ablation (w/o. normal) by removing both the normal prediction branch and normal map conditioning, reducing the model to RGB-only generation. Qualitative comparisons reveal that this simplified variant produces geometrically inconsistent deformations and structural artifacts, as shown in Fig.~\ref{fig:ablation}. Our complete framework, with joint appearance-geometry modeling, demonstrates superior geometric awareness and structural consistency while maintaining photorealistic quality.

\paragraph{Multi-Layer Geometry Fusion.} As shown in Tab.~\ref{tab:ablation_study}, without the guidance of geometry features from the MLGF module (w/o. MLGF), the model generates inaccurate poses even when appearance priors are provided. This occurs despite the photorealistic human rendering achieved, as evidenced by the lower FID and LPIPS scores. The qualitative results in Fig.~\ref{fig:ablation} further demonstrate that the model misinterprets clothing geometry, such as confusing the legs with the dress fabric, leading to incorrect human poses. This shows that our MLGF module injects superior geometry and missing texture cues, which greatly help resolve ambiguities and dramatically improve both pose accuracy and appearance quality.

\paragraph{Human NeRF.} We conduct an ablation study (w/o. Human NeRF) to evaluate the effectiveness of integrating coarse image conditioning signals produced by Human NeRF. When the diffusion model is conditioned without the coarse image from Human NeRF, the generation suffers from appearance ambiguities, such as hand artifacts (as shown in Fig.~\ref{fig:ablation}). This demonstrates the importance of Human NeRF in providing rich appearance priors that guide the generation process.

\subsection{Extensive Experimental Validation}  
We conducted a range of experiments to evaluate our method, including controlled and in-the-wild user studies as well as comparisons with state-of-the-art 3DGS-based methods such as IDOL~\cite{idol}, which rely on explicit 3D reconstruction for novel pose and view synthesis. Our method consistently outperforms existing approaches in perceptual quality and robustness, and demonstrates strong generalizability to real-world images from datasets such as DeepFashion~\cite{Deepfashion_dataset} and SHHQ~\cite{SHHQ_dataset}. Detailed results, qualitative comparisons, and further analyses are provided in the supplementary material.

\section{Limitation and Failure Cases}
Despite the strong performance demonstrated in our experiments, our method still faces several limitations and failure cases. It struggles to generate fine-grained details in small regions such as hands and facial features. Under extreme poses with severe self-occlusion (e.g., curled or crouched postures), large portions of clothing may become invisible in the reference image, leading to incomplete appearance recovery. In addition, our framework does not model physical dynamics such as loose clothing or hair motion. These limitations suggest promising directions for future improvements, such as incorporating higher-resolution modeling, physics-informed priors, or temporal dynamics.

\section{Conclusion}
We presented Blur2Sharp, a novel framework for high-fidelity human novel view and pose synthesis from a single image by combining the strengths of 3D-aware neural rendering and diffusion models. Our key innovation lies in our two-stage pipeline: a Human NeRF module ensures geometric consistency across poses and views, while a dual-domain diffusion model refines details using both RGB and normal map guidance. By integrating SMPL-based geometric priors (texture, normal, and semantic maps) through hierarchical feature fusion, our method achieves superior control over global structure and local appearance. Experiments demonstrate state-of-the-art performance in novel pose synthesis across multiple views, particularly for challenging cases like loose clothing.

\newpage
\section*{Acknowledgment}
This work was financially supported in part (project number: 112UA10019) by the Co-creation Platform of the Industry Academia Innovation School, NYCU, under the framework of the National Key Fields Industry-University Cooperation and Skilled Personnel Training Act, from the Ministry of Education (MOE) and industry partners in Taiwan. It also supported in part by the National Science and Technology Council, Taiwan, under Grant NSTC-114-2218-E-A49 -024, - Grant NSTC-112-2221-E-A49-089-MY3, Grant NSTC-114-2425-H-A49-001, Grant NSTC-113-2634-F-A49-007, Grant NSTC-112-2221-E-A49-092-MY3, and in part by the Higher Education Sprout Project of the National Yang Ming Chiao Tung University and the Ministry of Education (MOE), Taiwan. It was also partly supported by MediaTek Inc.; Hon Hai Research Institute; Gear Radio Electronics Corp.; E.SUN Commercial Bank, Ltd.; Penpower Technology Ltd.; and the Industrial Technology Research Institute.
{
    \small

}
\maketitlesupplementary


\section{Boarder Impacts} 
Our model might raise serious concerns regarding privacy and potential misuse. As these models can generate realistic and manipulable representations of individuals without their consent, they may be exploited to create non-consensual deepfakes, impersonate identities, or produce misleading visual content. This causes a high risk for public figures and vulnerable individuals because a single publicly available photo could be repurposed for malicious intent. We strongly encourage readers to use this work responsibly and strictly within the bounds of legal and ethical guidelines.


\section{Training Pipeline and Network Learning Details}
\label{supp:Implementation Details}
Our training pipeline consists of two main stages. Since we evaluate our method on two datasets, we train separate models for each dataset independently. Notably, the two datasets differ in camera viewpoints, which introduces a domain gap. In the first stage, we train the Human NeRF model for 40{,}000 iterations on the MVHumanNet~\cite{dataset_mvhumannet} dataset and 33{,}000 iterations on the HuMMan~\cite{dataset_humman} dataset using a batch size of 1. Input images are first center-cropped to a resolution of \(1024 \times 1024\) pixels, then resized and normalized to \(512 \times 512\). Once trained, the Human NeRF model generates coarse renderings for each sample, which are saved for use in the subsequent refinement stage. To filter out background regions from the coarse images, we apply a threshold of 0.5 to the human region mask predicted by the Human NeRF model.

The second stage of our method refines the coarse outputs using a generative diffusion model trained in two phases. To initialize the model, we adopt pre-trained weights from Stable Diffusion 1.5~\cite{latent_diffusion} for both the multi-view Denoising UNet and the Reference Network, while the 1D attention module is initialized with weights from AnimateDiff V3~\cite{guo2023animatediff}. In the first training phase, we train a base version of the diffusion model without multi-view and 1D attention for 80{,}000 iterations on MVHumanNet and 50{,}000 iterations on HuMMan, using a learning rate of \(1 \times 10^{-5}\) and a batch size of 2. In the second phase, we freeze the MLGF module, the RGB and normal encoders, and the Reference Network. We then enable both multi-view and 1D attention mechanisms, and fine-tune only the multi-view Denoising UNet for 30{,}000 iterations on MVHumanNet and 15{,}000 iterations on HuMMan. The entire training process takes approximately five days on a single NVIDIA A6000 GPU.


\section{Training Objective} For the Human NeRF training stage, following ~\cite{sherf}, we optimize a composite objective that integrates both photometric and geometric constraints. The primary reconstruction loss \( \mathcal{L}_{\text{recon}} \) measures pixel-level discrepancies between the rendered and ground truth images. To ensure silhouette consistency, we include a mask loss \( \mathcal{L}_{\text{mask}} \), implemented as the binary cross-entropy between predicted and ground truth masks. To further enhance perceptual quality, we incorporate a structural similarity loss \( \mathcal{L}_{\text{SSIM}} \) and a perceptual feature loss \( \mathcal{L}_{\text{LPIPS}} \), computed using a pretrained VGG network. The complete objective for Human NeRF training is defined as:

\begin{equation}
\mathcal{L}_{\text{NeRF}} = \mathcal{L}_{\text{recon}} + \lambda_{\text{mask}} \mathcal{L}_{\text{mask}} + \lambda_{\text{SSIM}} \mathcal{L}_{\text{SSIM}} + \lambda_{\text{LPIPS}} \mathcal{L}_{\text{LPIPS}},
\end{equation}

where $\lambda_{\text{mask}}$, $\mathcal{L}_{\text{mask}}$, and $\mathcal{L}_{\text{SSIM}}$ are loss weights. 

The diffusion refinement stage employs a dual-domain velocity-based prediction (v-prediction) objective~\cite{salimans2022progressive} that jointly optimizes RGB appearance and surface normal geometry. The RGB branch is conditioned on three inputs: the reference image \( I_{\text{ref}} \), RGB features \( F_{\text{rgb}} \) derived from Human NeRF renderings, and geometry features \( F_{\text{geo}} \), which provide structural pose guidance. Similarly, the normal branch is conditioned on the reference normal map \( N_{\text{ref}} \), normal features \( F_{\text{normal}} \), and the shared geometry features \( F_{\text{geo}} \). The v-prediction objective is formally defined as:

\begin{equation}
\begin{split}
\mathcal{L}_{\text{v-pred}}
  = \mathbb{E}_{\cdot} \Big[
    \left\| \hat{v}_\theta \big(I_t; I_{\text{ref}}, F_{\text{rgb}}, F_{\text{geo}}\big) - v_t^I \right\|_2^2 \\
    +\; \left\| \hat{v}_\theta \big(N_t; N_{\text{ref}}, F_{\text{normal}}, F_{\text{geo}}\big) - v_t^N \right\|_2^2
  \Big].
\end{split}
\end{equation}

where \( I_t = \alpha_t I + \sigma_t \epsilon_I \) and \( N_t = \alpha_t N + \sigma_t \epsilon_N \) denote the noisy RGB and normal latents at timestep \( t \), respectively. \( v_t^I \) and \( v_t^N \) represent the target velocities for the RGB and normal branches.


\section{Supplementary Network Architecture Specifications}
\label{supp:Additional Implementation Details}
\begin{figure*}[]
  \includegraphics[scale=0.5]{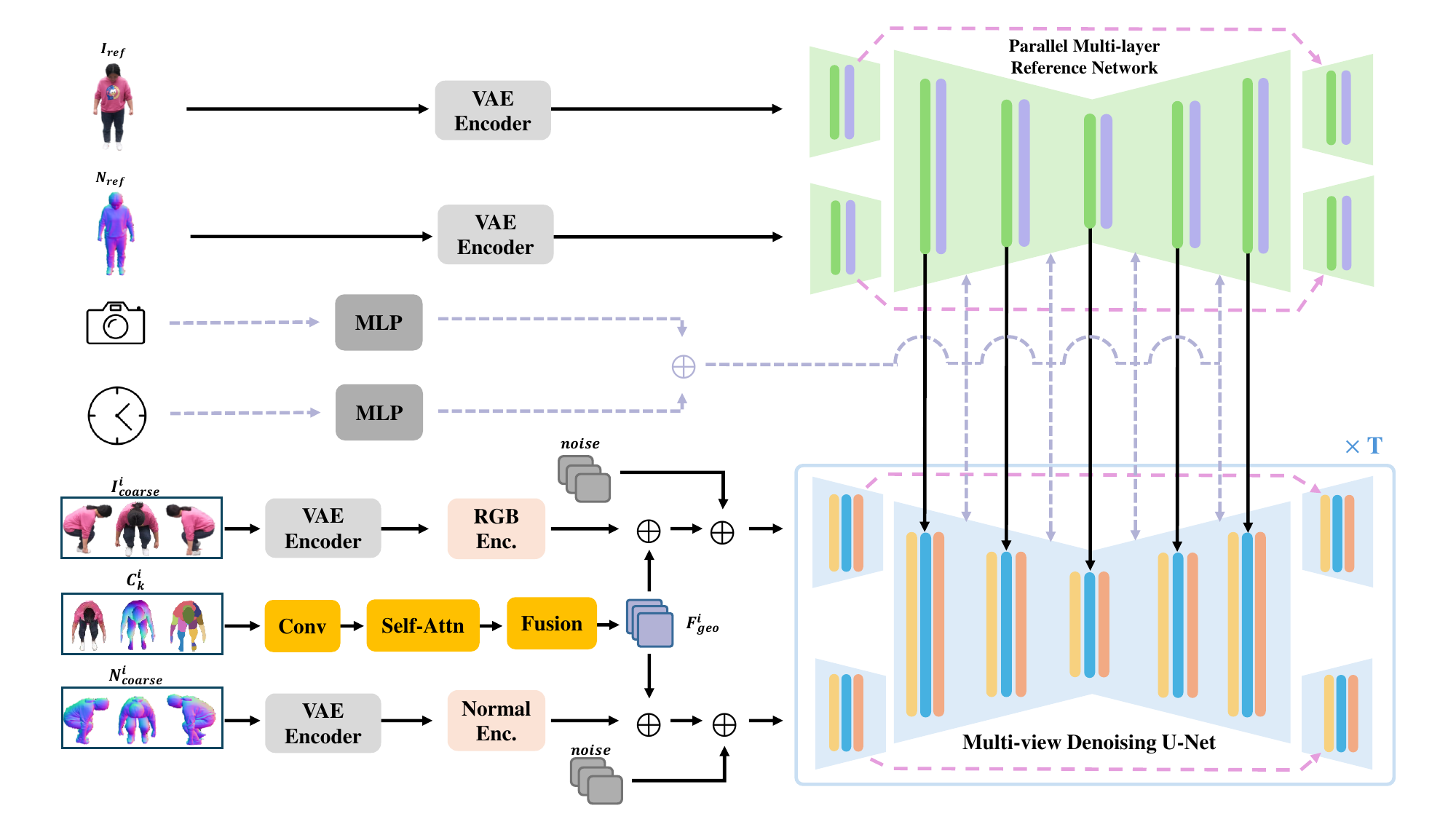}  
  \centering
  \caption{Architecture of Generative Prior Refinement. The network processes coarse NeRF renderings $I_{\text{coarse}}$ and predicted normal maps $N_{\text{coarse}}$ through RGB encoder and normal encoder. Extracted features are combined with geometry features $F_{\text{geo}}$ via element-wise addition ($\oplus$) before injection into our multi-view diffusion model. Note that \( \mathbf{C}_k^i \) denotes the three condition maps (i.e., texture, normal, and semantic) used in MLGF module.}
  \label{fig:archictecture}  
\end{figure*}

\paragraph{Details of Coarse Image Conditioning.} In Stage 1, We utilize SHERF~\cite{sherf} to generate coarse renderings $I_{\text{coarse}}$ from a single reference image, capturing both novel poses and viewpoints. These initial renderings preserve the subject's appearance while approximating the target geometry. We then predict the corresponding normal maps $N_{\text{coarse}}$ using a pretrained off-the-shelf normal estimator~\cite{khirodkar2024sapiens}. For conditioning the generative model, both the coarse RGB images and normal maps are encoded into latent representations using VAE and then processed through separate encoder networks $E_{\text{rgb}}$ and $E_{\text{normal}}$ as shown in Fig.\ref{fig:archictecture}, each network is consisting of four convolutional layers as structured in Tab.~\ref{rgb_encoder}. The resulting latent features are combined with geometry features $F_{\text{geo}}$ from MLGF via element-wise addition, followed by convolutional processing of the noise distributions in both the RGB and normal domains.

\begin{table}[]
\centering
\begin{tabular}{|c|c|}
\hline
\textbf{Architecture} & \textbf{Channels} \\
\hline
Conv2d (kernel: 3$\times$3, stride: 1) & 4 $\rightarrow$ 16 \\
SiLU & - \\
\hline
Conv2d (kernel: 3$\times$3, stride: 1) & 16 $\rightarrow$ 16 \\
SiLU & - \\
Conv2d (kernel: 3$\times$3, stride: 1) & 16 $\rightarrow$ 32 \\
SiLU & - \\
\hline
Conv2d (kernel: 3$\times$3, stride: 1) & 32 $\rightarrow$ 32 \\
SiLU & - \\
Conv2d (kernel: 3$\times$3, stride: 1) & 32 $\rightarrow$ 96 \\
SiLU & - \\
\hline
Conv2d (kernel: 3$\times$3, stride: 1) & 96 $\rightarrow$ 96 \\
SiLU & - \\
Conv2d (kernel: 3$\times$3, stride: 1) & 96 $\rightarrow$ 256 \\
SiLU & - \\
\hline
Conv2d (kernel: 3$\times$3, stride: 1), zero-initialized & 256 $\rightarrow$ 320 \\
\hline
\end{tabular}
\caption{Architecture of RGB and normal encoders}
\label{rgb_encoder}
\end{table}

\paragraph{Conditioning on Camera View and Time Step.}
Unlike prior methods such as MagicMan~\cite{he2024magicman} that only leverage camera rotations for viewpoint conditioning, we incorporate both the rotation matrix $\mathbf{R} \in \mathbb{R}^{3 \times 3}$ and the translation vector $\mathbf{t} \in \mathbb{R}^{3 \times 1}$ to facilitate more precise viewpoint control. These extrinsic parameters together define the full camera-to-world transformation. We flatten the rotation and translation matrix into a 12-dimensional vector $\mathbf{v} \in \mathbb{R}^{12}$ and pass it through a feed-forward network to obtain the camera embeddings $\mathbf{f}_{\text{cam}} \in \mathbb{R}^{1024}$. These embeddings are then combined with the denoising time step embeddings $\mathbf{f}_{\text{time}} \in \mathbb{R}^{1024}$ of the diffusion model via element-wise addition to provide viewpoint-aware conditioning during the diffusion process. As shown in Tab.~\ref{tab:ablation_study_camera}, this design leads to consistently better results compared to using rotations alone.

\begin{figure}[]
  \includegraphics[scale=0.4]{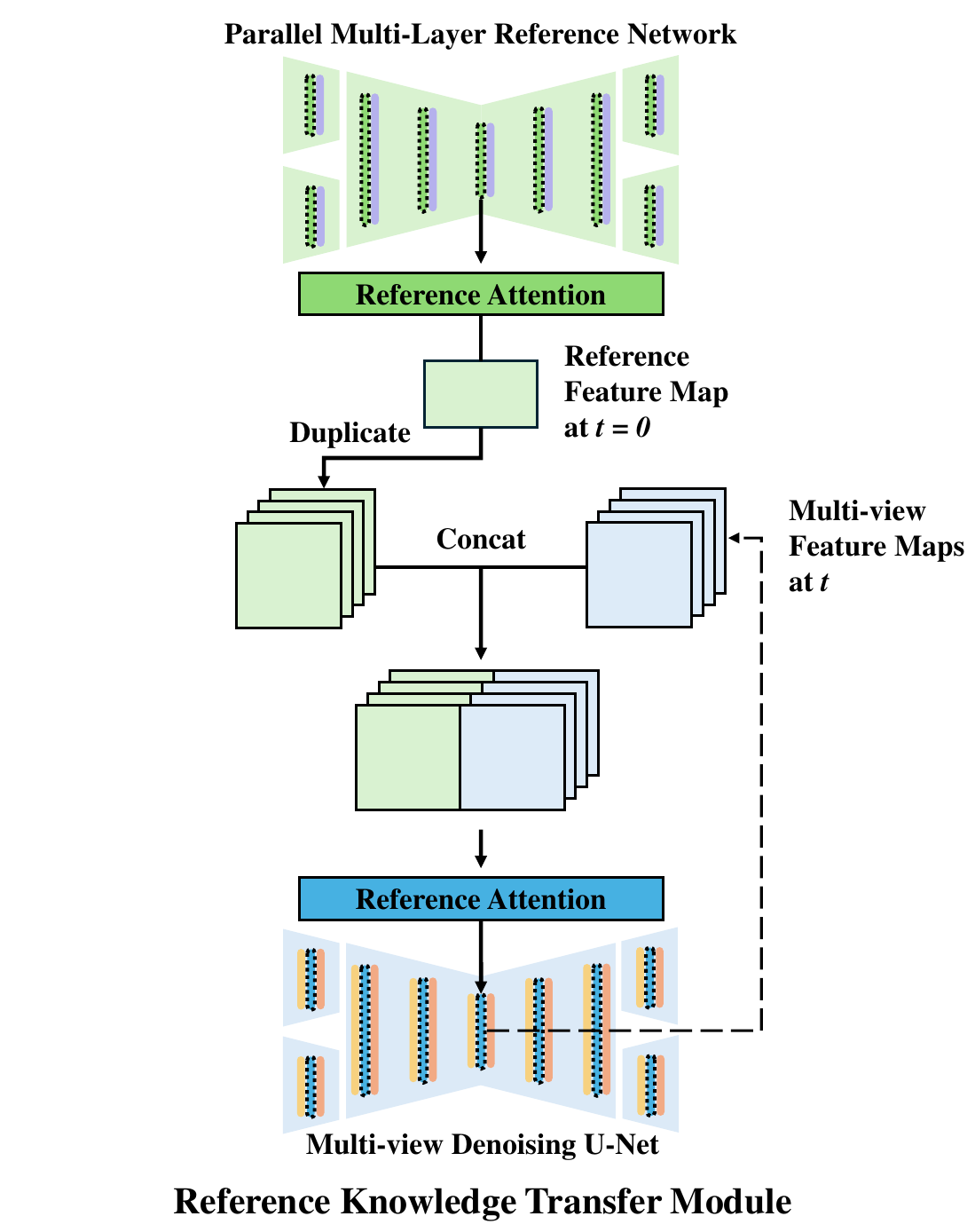}  
  \centering
  \caption{Architecture of Reference Knowledge Transfer Module. The reference feature map is extracted via the reference attention mechanism in the Parallel Multi-Layer Reference Network and duplicated to match the dimension of the multi-view feature map. The duplicated reference feature map is then concatenated with the multi-view feature maps and passed into the reference attention of the Multi-view Denoising UNet for knowledge transfer.}
  \label{fig:reference_knowledge}  
\end{figure}

\paragraph{Details of Reference Knowledge Transfer Module.} Similar to MagicMan~\cite{he2024magicman}, We implement a Reference Knowledge Transfer Module to effectively propagate informative features from a single reference view to all target views, thereby enhancing appearance consistency between the generated multi-view images and the reference image. As illustrated in Fig.~\ref{fig:reference_knowledge}, the module first extracts a reference feature map from the reference attention block of the Reference Network. This feature map encodes high-level appearance information from the reference image. To enable alignment across views, the reference feature map is spatially and dimensionally duplicated to match the shape of the multi-view feature maps. These duplicated reference features are then concatenated with the multi-view features and passed to the reference attention layers of the denoising UNet, where attention computations integrate reference-guided appearance cues into the multi-view generation process.

\paragraph{Details of Dual-Branch RGB-Normal Conditional Diffusion Model.} As shown in Fig.\ref{fig:archictecture}, to enable the joint generation of RGB images and normal maps, we first extract features separately from the RGB and normal latents using their respective downsampling blocks. The resulting intermediate features are then stacked and averaged to form a unified representation that captures both appearance and geometric information. This unified representation is passed through the shared intermediate layers of the UNet, while the residuals from the early modality-specific blocks are preserved and routed through their corresponding upsampling layers using the skip connections. This design ensures that each modality retains its unique characteristics throughout the denoising process.


\section{Applications}
\begin{figure*}[]
  \includegraphics[scale=0.45]{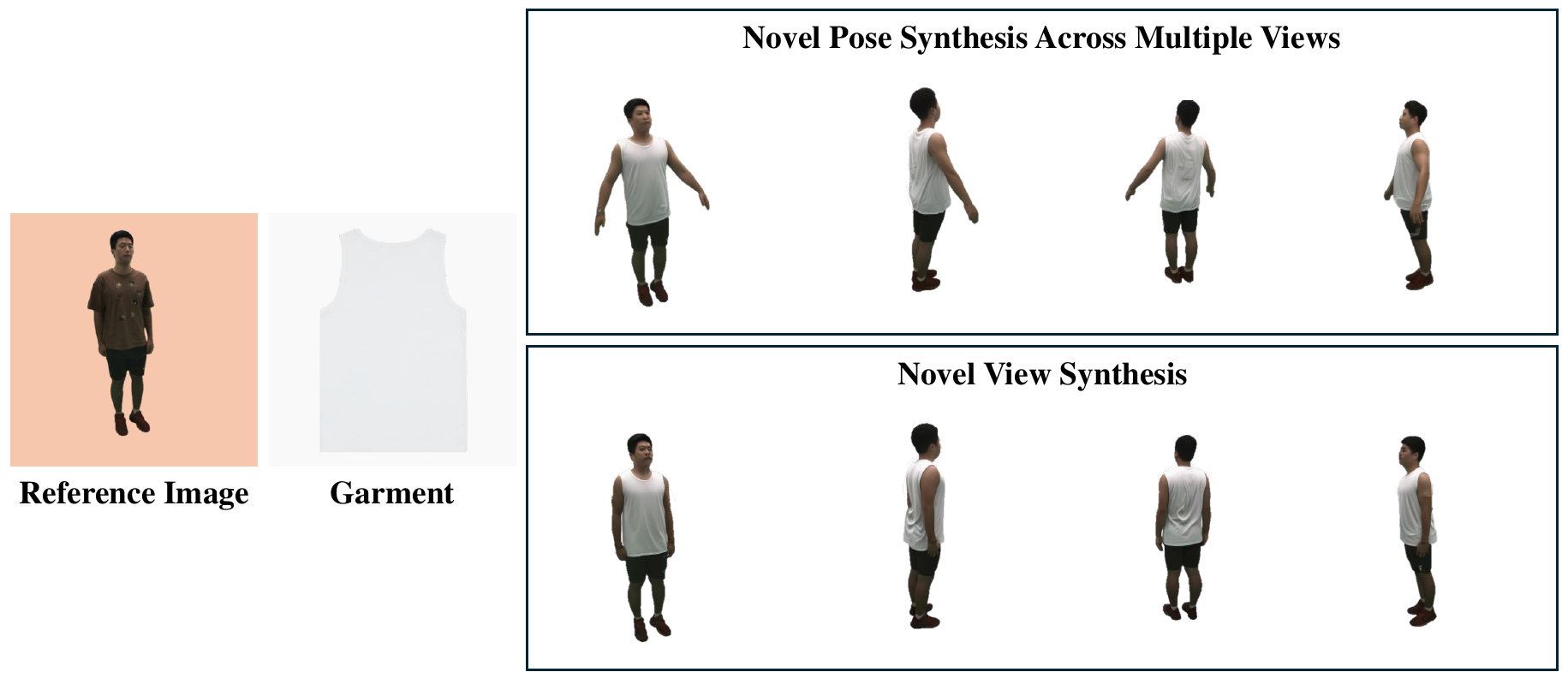}  
  \centering
    \caption{Given a reference image and a target garment, we first employ a 2D virtual try-on model to generate a try-on image. Blur2Sharp then synthesizes multi-view images under novel poses and viewpoints based on this try-on result.}
  \label{fig: virtual_try_on}  
\end{figure*}

Our method can be directly applied to realistic multi-view and multi-pose virtual try-on, enabling photorealistic visualization of how a garment appears from various angles and body poses using only a single reference image. Given a reference image of a person and a target garment, we first apply a pretrained 2D virtual try-on model~\cite{choi2024improving} to generate a 2D try-on result aligned with the reference pose. We then apply \textbf{Blur2Sharp} to produce high-quality images under both novel poses and novel viewpoints, conditioned on the 2D try-on image. As shown in Fig.~\ref{fig: virtual_try_on}, \textbf{Blur2Sharp} is capable of generating realistic and consistent try-on results across diverse viewpoints and human poses.

\subsection{Additional Ablation Studies}
\paragraph{Evaluating Coarse RGB-Normal Conditioning strategies.}To better understand the effectiveness of our proposed dual-branch conditioning mechanism, we conduct an ablation study comparing different methods of injecting coarse RGB renderings and predicted normal maps into the diffusion model. As an alternative baseline, we experiment with a simple conditioning approach in which the RGB and normal latents are concatenated with the noise input along the channel dimension. To accommodate this change, we modify the input convolutional block of the diffusion network to accept an 8-channel input and initialize the weights accordingly.

\begin{figure*}[]
  \includegraphics[scale=0.65]{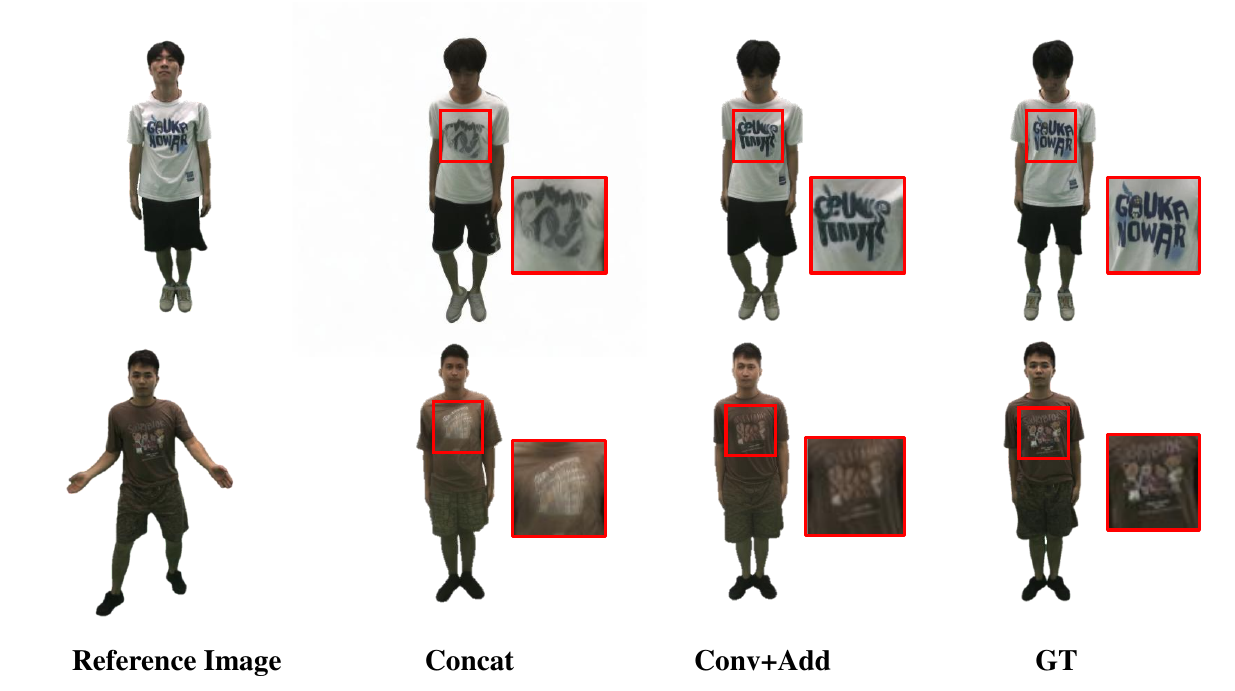}  
  \centering
    \caption{Ablation studies and qualitative comparisons of different RGB-Normal conditioning strategies. \textcolor{red}{Red} boxes indicate enlarged regions.}
  \label{fig:ablation_concat}  
\end{figure*}

As shown in Fig.~\ref{fig:ablation_concat}, this naive concatenation strategy fails to recover high-frequency texture details, despite maintaining accurate pose geometry. The approach yields significantly worse perceptual quality metrics, with higher LPIPS and FID scores reported in Tab.~\ref{tab:RGB-Normal Conditioning Strategies}. We hypothesize that the simple concatenation mechanism indiscriminately mixes all conditioning signals—including blurry features—without effective feature decomposition or selective enhancement, allowing low-frequency artifacts from the input RGB and normal maps to persist throughout the denoising process. 

\begin{table}[]
\centering
\caption{Quantitative comparison of different coarse RGB-Normal conditioning strategies for novel pose and view synthesis.}
\resizebox{\columnwidth}{!}{%
\begin{tabular}{l|cccc|cccc}
\toprule
Method & \multicolumn{4}{c|}{Novel Pose Synthesis} & \multicolumn{4}{c}{Novel View Synthesis} \\

 & PSNR $ \uparrow$ & SSIM $ \uparrow$ & LPIPS $ \downarrow$ & FID $ \downarrow$ & PSNR $ \uparrow$ & SSIM $ \uparrow$ & LPIPS $ \downarrow$ & FID $ \downarrow$ \\
\midrule
Concat & 22.31 & 0.940 & 0.047 & 34.04 & 22.56 & 0.942 & 0.046 & 34.04\\
Conv+Add & \textbf{23.31} & \textbf{0.946} & \textbf{0.039} & \textbf{24.38} & \textbf{23.82} & \textbf{0.948} & \textbf{0.036} & \textbf{23.50} \\
\bottomrule
\end{tabular}%
}
\label{tab:RGB-Normal Conditioning Strategies}
\end{table}

\paragraph{Evaluating Different Inputs of Camera Conditioning.}
To assess the contribution of our camera conditioning approach, we perform an ablation study by removing the translation component from the extrinsic matrix and conditioning the model using only the 3×3 rotation matrix. Our full method uses the complete 3×4 extrinsic matrix, which includes both rotation and translation. Quantitative results in Tab.~\ref{tab:ablation_study_camera} show that removing translation leads to a drop in performance across all metrics, indicating that the translation matrix provides valuable information for improving generation quality.

\begin{table*}[]
\centering
\caption{Quantitative ablation study of camera inputs on MVHumanNet dataset.}
\resizebox{\textwidth}{!}{%
\begin{tabular}{l|cccc|cccc}
\toprule
Method & \multicolumn{4}{c|}{Novel Pose Synthesis} & \multicolumn{4}{c}{Novel View Synthesis} \\

 & PSNR $ \uparrow$ & SSIM $ \uparrow$ & LPIPS $ \downarrow$ & FID $ \downarrow$ & PSNR $ \uparrow$ & SSIM $ \uparrow$ & LPIPS $ \downarrow$ & FID $ \downarrow$ \\

\midrule
Rotation only & 23.27 & 0.946 & 0.039  & 25.21 & 23.79 & 0.948 & 0.036 & 24.79 \\
Rotation and Translation & \textbf{23.31} & \textbf{0.946} & \textbf{0.039} & \textbf{24.38} & \textbf{23.82} & \textbf{0.948} & \textbf{0.036} & \textbf{23.50}  \\
\bottomrule
\end{tabular}%
}
\label{tab:ablation_study_camera}
\end{table*}

\begin{figure*}[]
  \includegraphics[scale=0.7]{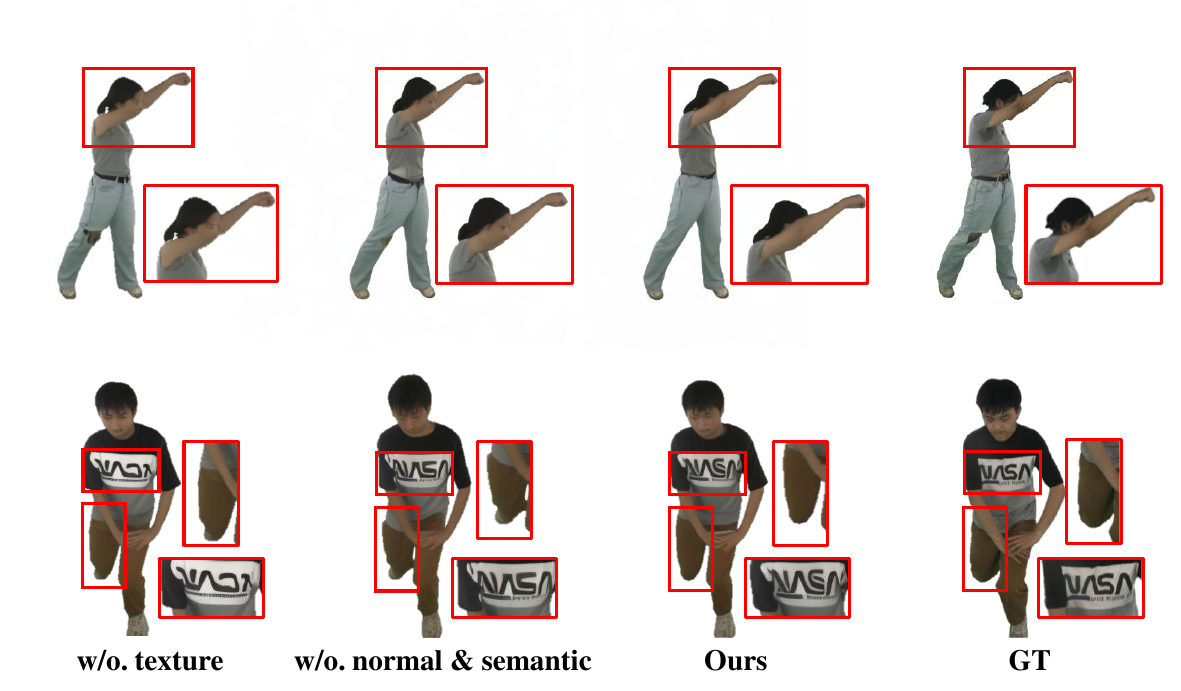}  
  \centering
    \caption{Qualitative results of different SMPL Conditions. \textcolor{red}{Red} boxes indicate enlarged regions.}
  \label{fig:ablation_SMPL}  
\end{figure*}

\paragraph{Evaluating Different SMPL Conditions.}As shown in Tab.~\ref{tab:ablation_study_SMPL}, incorporating all three SMPL priors (texture, normal, and semantic maps) leads to notable improvements in both structural alignment and perceptual quality for novel pose and view synthesis. Specifically, the inclusion of SMPL normal and semantic maps (w/o. texture) enhances structural alignment (higher PSNR) across views and poses by providing geometric cues, while the variant model with SMPL texture priors only (w/o. normal \& semantic) contributes to improved perceptual realism (lower FID) by preserving fine-grained surface appearance. Our full model achieves the lowest FID and reduced LPIPS, indicating enhanced visual fidelity, while maintaining strong PSNR and SSIM values that reflect better structural consistency compared to ablated variants. We provide the qualitative results in Fig.~\ref{fig:ablation_SMPL}.

\begin{table*}[!htbp]
\centering
\caption{Quantitative ablation study on different input configurations of the MLGF module. "w/o. texture" denotes the model variant without SMPL texture maps, while "w/o. normal \& semantic" refers to the configuration excluding both SMPL normal and semantic maps. "Ours" represents the proposed method that integrates SMPL texture, normal, and semantic priors for comprehensive geometric guidance. The top two methods are highlighted in \textbf{bold} and \underline{underline}.}
\resizebox{\textwidth}{!}{%
\begin{tabular}{l|cccc|cccc}
\toprule
Method & \multicolumn{4}{c|}{Novel Pose Synthesis} & \multicolumn{4}{c}{Novel View Synthesis} \\

 & PSNR $ \uparrow$ & SSIM $ \uparrow$ & LPIPS $ \downarrow$ & FID $ \downarrow$ & PSNR $ \uparrow$ & SSIM $ \uparrow$ & LPIPS $ \downarrow$ & FID $ \downarrow$ \\

\midrule
w/o. texture & \underline{23.28} & 0.945 & 0.039  & 28.08 & \underline{23.81} & \underline{0.948} & 0.037 & 27.81 \\
w/o. normal \& semantic & 23.18 & \underline{0.945} & \underline{0.039} & \underline{27.44} & 23.63 & 0.947 & \underline{0.037} & \underline{26.98}  \\
\textbf{Ours} & \textbf{23.31} & \textbf{0.946} & \textbf{0.039} & \textbf{24.38} & \textbf{23.82} & \textbf{0.948} & \textbf{0.036} & \textbf{23.50} \\
\bottomrule
\end{tabular}
}
\label{tab:ablation_study_SMPL}
\end{table*}

\subsection{Efficiency Analysis}
Tab.~\ref{tab:efficiency_analysis} presents the memory usage and computation time required for each stage of the pipeline. 
\begin{table}[h!]
\centering
\caption{Memory and Runtime Efficiency of each stage of our proposed method. }
\label{tab:efficiency_analysis}
\resizebox{\columnwidth}{!}{%
\begin{tabular}{|l|c|c|}
\hline
Method                      & Memory (GB) & Time (sec) \\
\hline
Rendering SMPL Condition Maps               & 2.47                 & 1.58                 \\
Human NeRF                         & 6.38                 & 4.51                 \\
Generative Prior Refinement                 & 6.87                 & 14.12                \\
\hline
\end{tabular}
}
\end{table}


\subsection{Additional Experimental Results}

\begin{figure*}[]
  \includegraphics[scale=0.5]{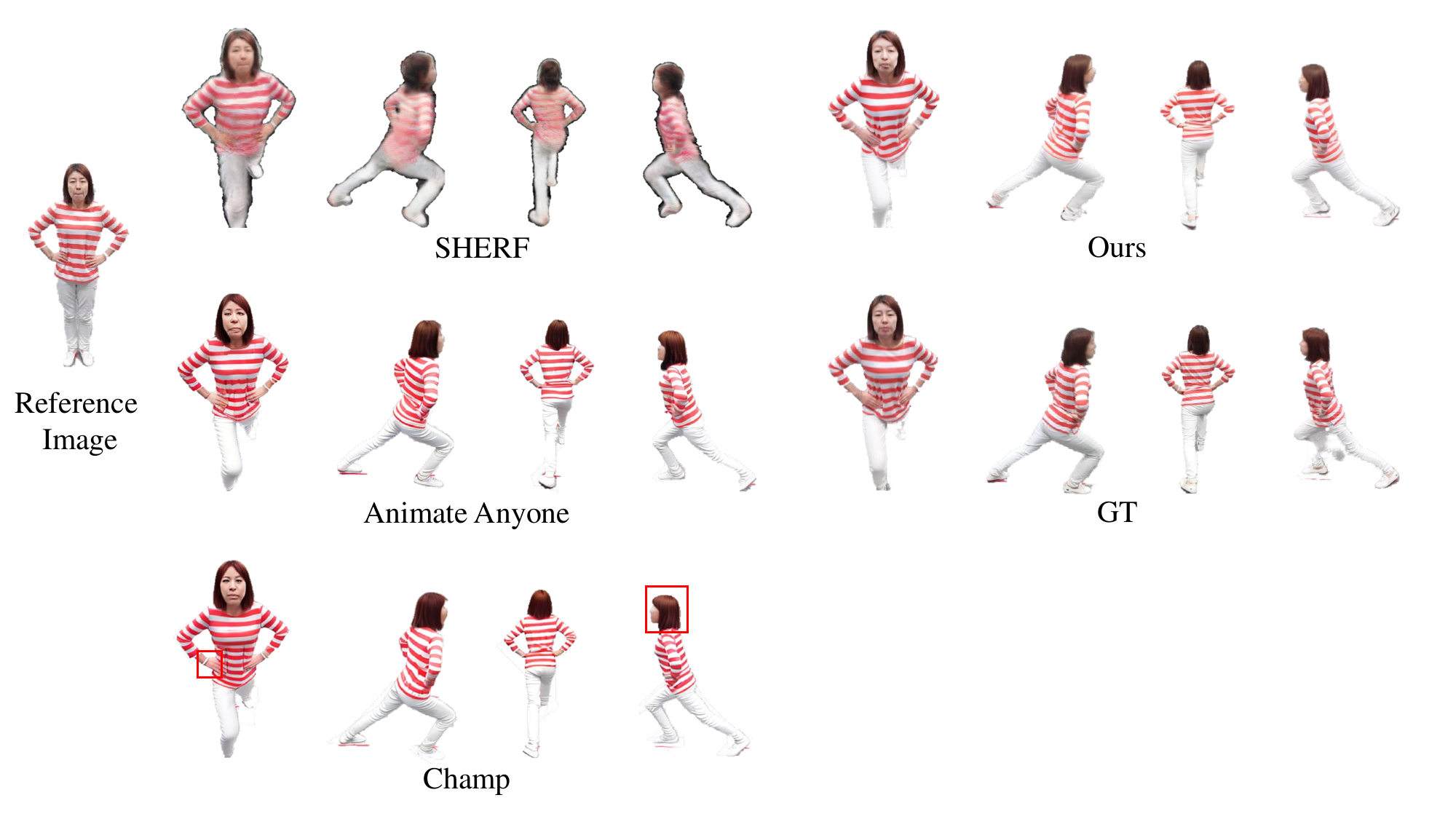}  
  \centering
    \caption{Qualitative results of novel pose synthesis across multiple views on the HuMMan~\cite{dataset_humman} dataset. We compare our method with SHERF~\cite{sherf}, Animate Anyone~\cite{AnimateAnyone}, and Champ~\cite{Champ}. \textcolor{red}{Red} boxes highlight appearance ambiguities or multi-view inconsistencies.}
  \label{fig:extra_result_1}  
\end{figure*}

\begin{figure*}[]
  \includegraphics[scale=0.5]{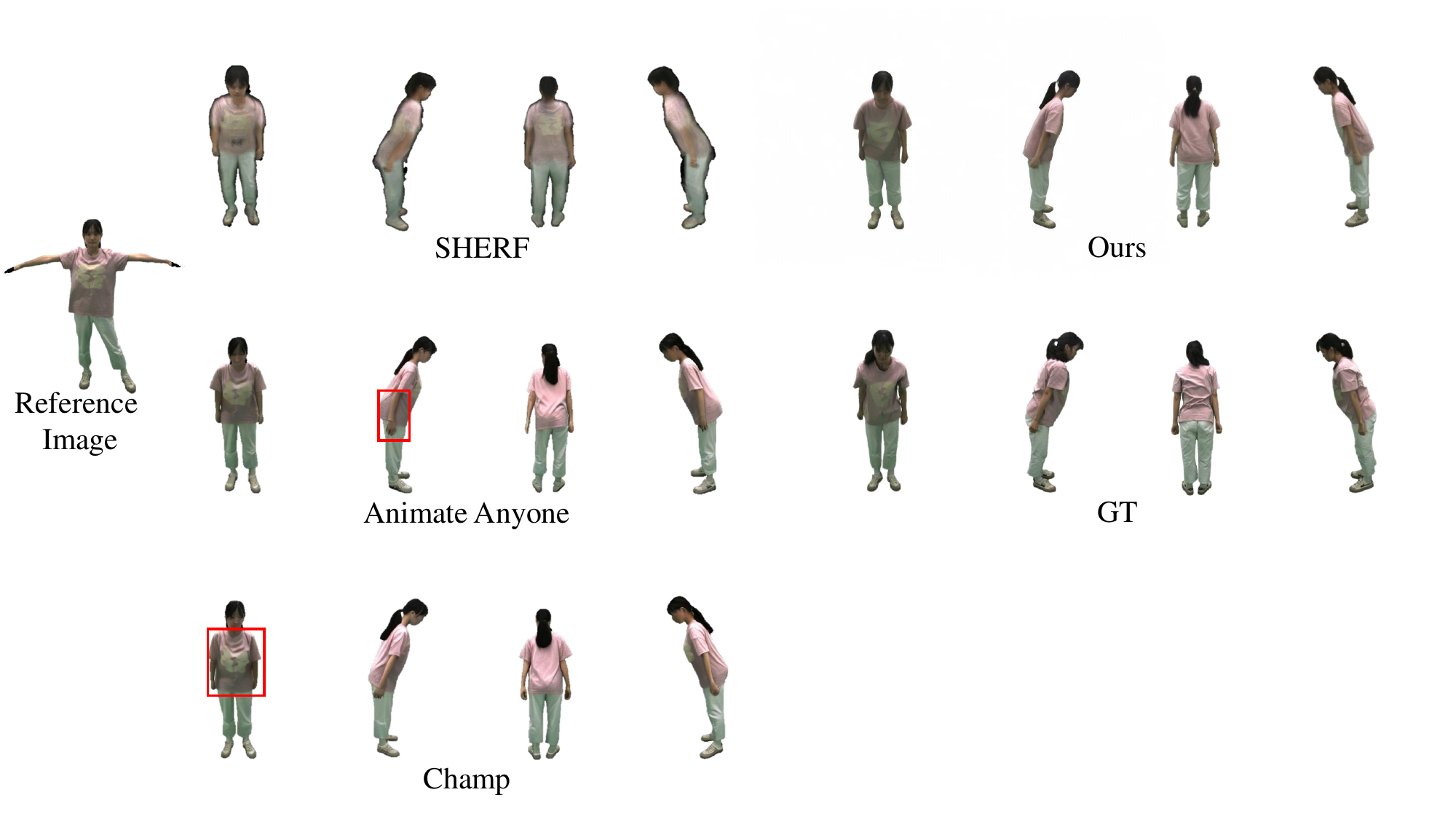}  
  \centering
    \caption{Qualitative results of novel pose synthesis across multiple views on MVHumanNet~\cite{dataset_mvhumannet} dataset. We compare our method with SHERF~\cite{sherf}, Animate Anyone~\cite{AnimateAnyone}, and Champ~\cite{Champ}. \textcolor{red}{Red} boxes highlight appearance ambiguities or multi-view inconsistencies.}
  \label{fig:extra_result_2}  
\end{figure*}

\begin{figure*}[]
  \includegraphics[scale=0.5]{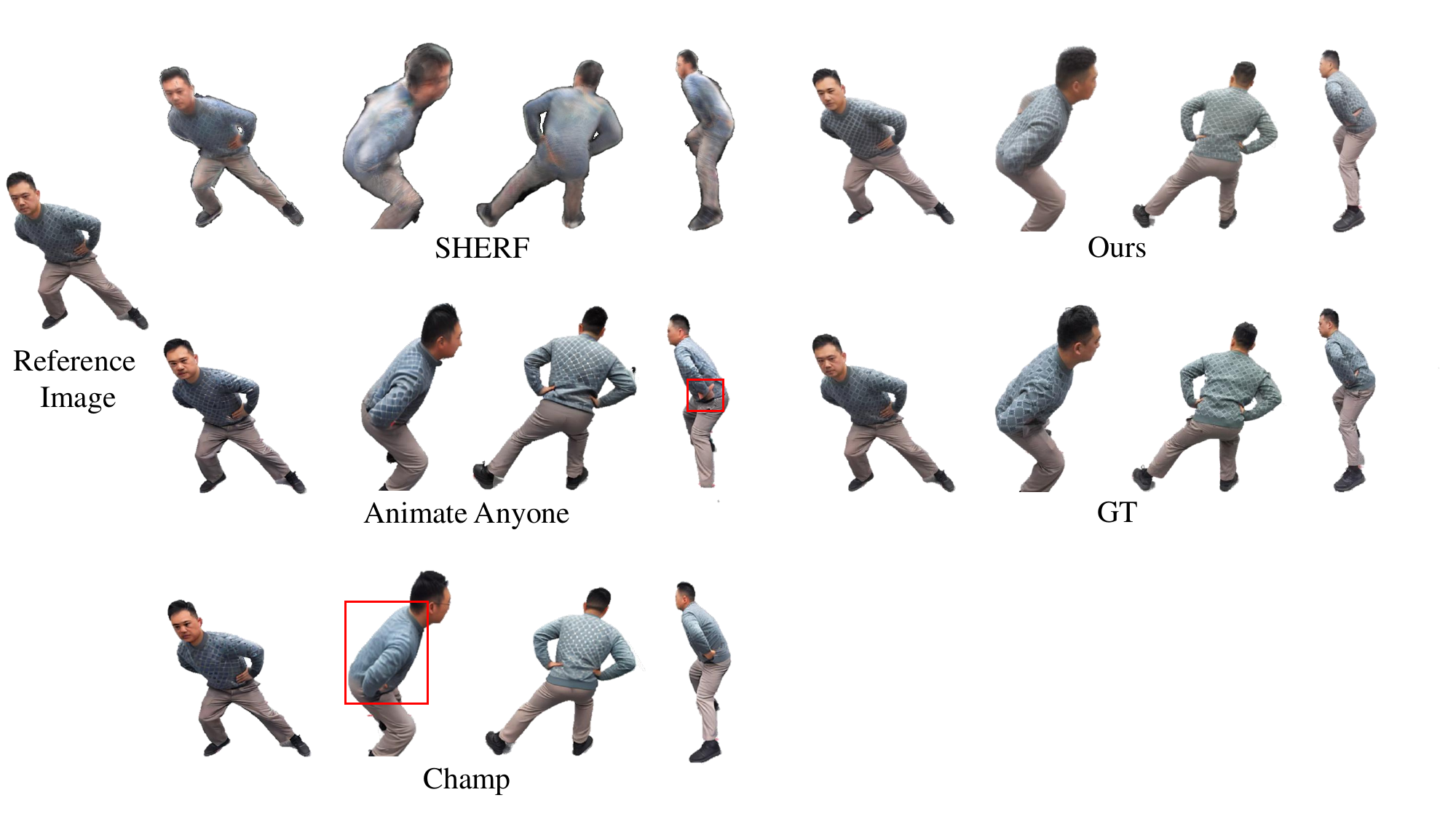}  
  \centering
    \caption{Qualitative results of novel view synthesis on HuMMan~\cite{dataset_humman} dataset. We compare our method with SHERF~\cite{sherf}, Animate Anyone~\cite{AnimateAnyone}, and Champ~\cite{Champ}. \textcolor{red}{Red} boxes highlight appearance ambiguities or multi-view inconsistencies.}
  \label{fig:extra_result_3}  
\end{figure*}

\begin{figure*}[]
  \includegraphics[scale=0.5]{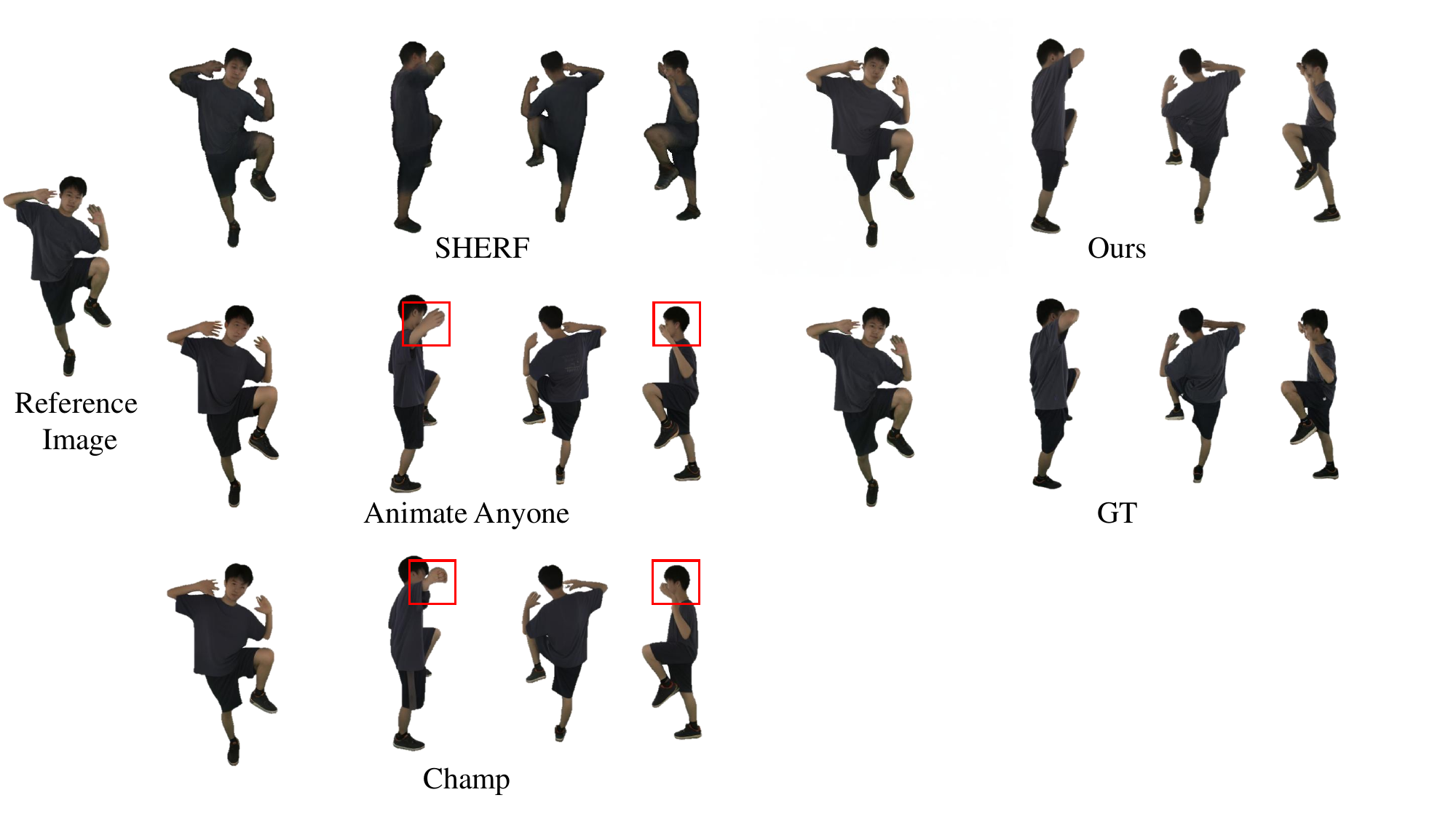}  
  \centering
    \caption{Qualitative results of novel view synthesis on MVHumanNet~\cite{dataset_mvhumannet} dataset. We compare our method with SHERF~\cite{sherf}, Animate Anyone~\cite{AnimateAnyone}, and Champ~\cite{Champ}. \textcolor{red}{Red} boxes highlight appearance ambiguities.}
  \label{fig:extra_result_4}  
\end{figure*}

\paragraph{Novel Pose Synthesis Across Multiple Views.} Fig.~\ref{fig:extra_result_1} and \ref{fig:extra_result_2} present additional qualitative results for novel pose synthesis across multiple views in comparison with state-of-the-art approaches. 
\paragraph{Novel View Synthesis}
Fig.~\ref{fig:extra_result_3} and \ref{fig:extra_result_4} present additional qualitative results for novel view synthesis in comparison with state-of-the-art approaches.

Our method demonstrates superior performance, exhibiting enhanced multi-view consistency and photorealism.


\paragraph{Comparison with 3DGS-based Methods.} 
To further investigate, we examined IDOL~\cite{idol}, a state-of-the-art 3DGS-based~\cite{gaussian_splatting} method that constructs an explicit 3D reconstruction to generate novel poses and views. We trained our purely 2D method on a subset of the HuGe100K dataset~\cite{idol}, which was originally introduced by IDOL~\cite{idol}, to evaluate its performance under similar conditions. As shown in Tab.~\ref{tab:3dgs_comparison}, our method achieves competitive perceptual quality without relying on explicit 3D reconstruction. While IDOL~\cite{idol} generally preserves the reference image’s geometry and appearance, even small deviations in pose or viewpoint can lead to floating artifacts, a common limitation of 3DGS-based approaches. Notably, these methods were developed around the same period, demonstrating that high-quality novel pose and view synthesis can be achieved using a simpler 2D approach. 

\begin{figure*}[]
  \includegraphics[scale=0.7]{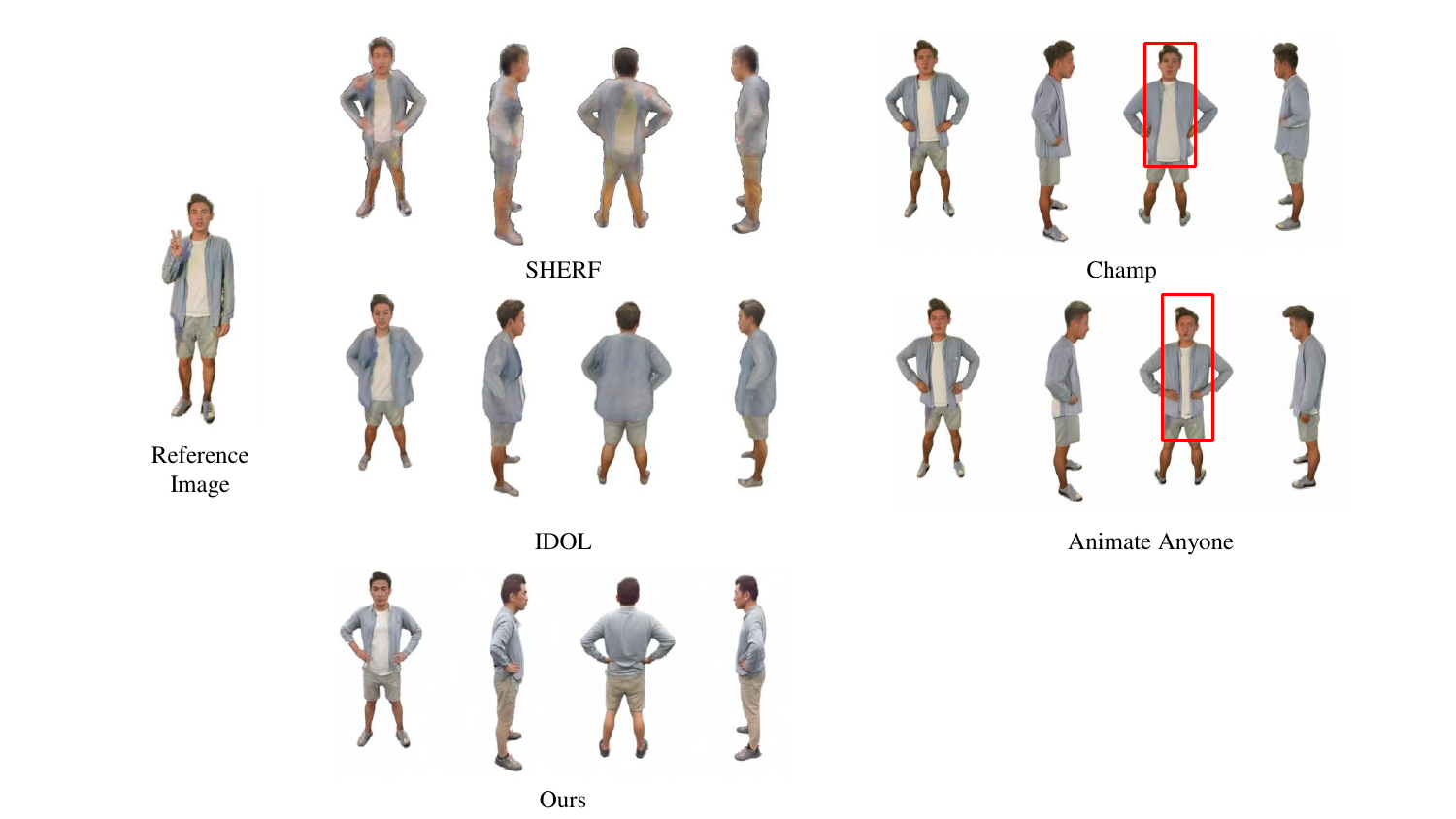}  
  \centering
    \caption{Qualitative results of in-the-wild setting across multiple views on the HuGe100K~\cite{idol} dataset. We compare our method with IDOL~\cite{idol}. \textcolor{red}{Red} boxes highlight appearance ambiguities or multi-view inconsistencies.}
  \label{fig:in_the_wild_1}  
\end{figure*}

\begin{figure*}[]
  \includegraphics[scale=0.7]{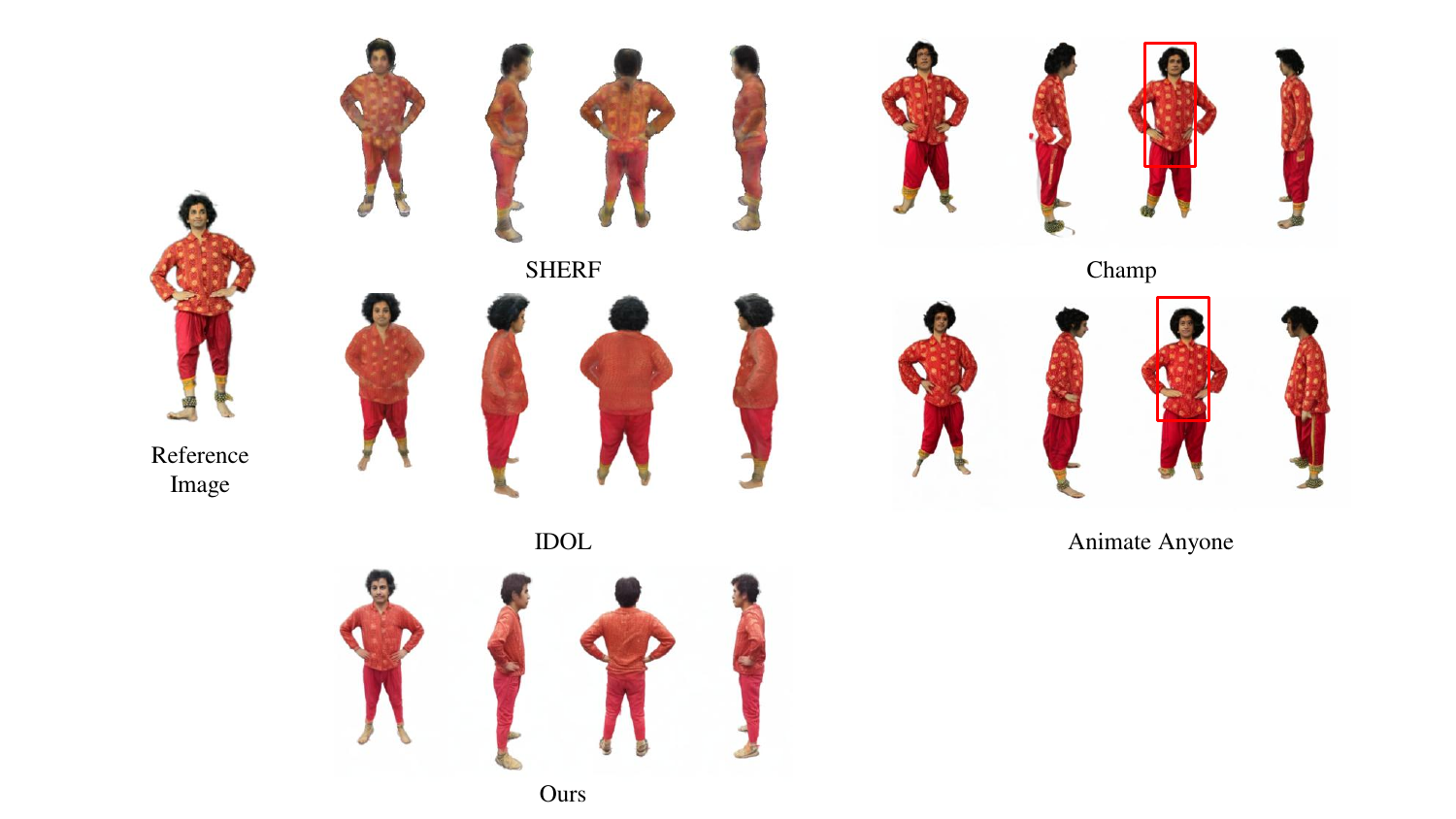}  
  \centering
    \caption{Qualitative results of in-the-wild setting across multiple views on the HuGe100K~\cite{idol} dataset. We compare our method with IDOL~\cite{idol}. \textcolor{red}{Red} boxes highlight appearance ambiguities or multi-view inconsistencies.}
  \label{fig:in_the_wild_2}  
\end{figure*}

\begin{figure*}[]
  \includegraphics[scale=0.7]{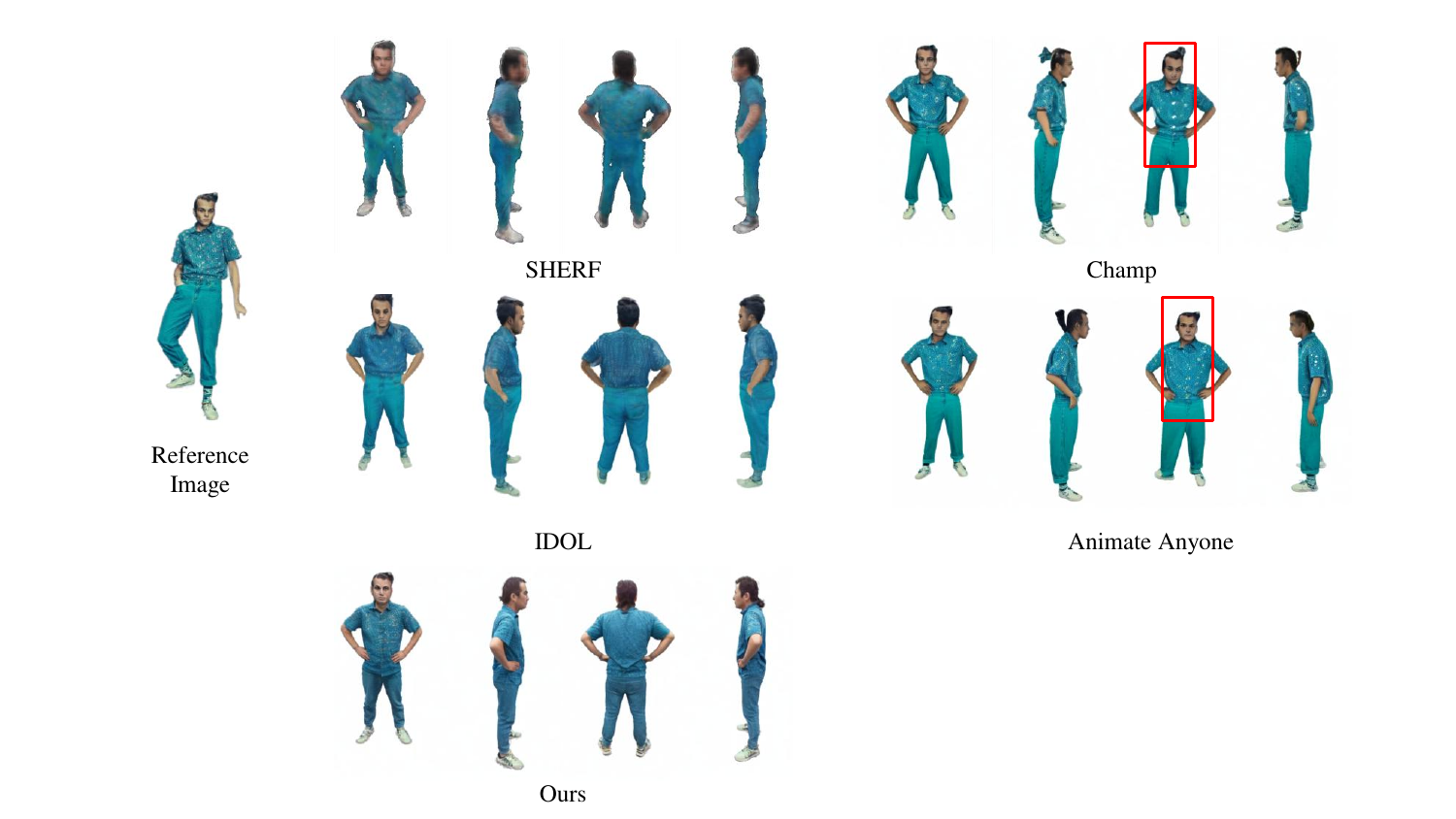}  
  \centering
    \caption{Qualitative results of in-the-wild setting across multiple views on the HuGe100K~\cite{idol} dataset. We compare our method with IDOL~\cite{idol}. \textcolor{red}{Red} boxes highlight appearance ambiguities or multi-view inconsistencies.}
  \label{fig:in_the_wild_3}  
\end{figure*}

\paragraph{Generalizability Analysis.}
To evaluate the generalizability of our method beyond controlled settings, we compare it with SHERF~\cite{sherf}, Animate Anyone~\cite{AnimateAnyone}, Champ~\cite{Champ}, and IDOL~\cite{idol} on in-the-wild datasets, DeepFashion~\cite{Deepfashion_dataset} and SHHQ~\cite{SHHQ_dataset}, using camera poses and SMPL parameters estimated by CLIFF \cite{li2022cliff}. All methods are trained on the HuMMan dataset~\cite{dataset_humman}, except for IDOL~\cite{idol}, which uses its official pre-trained weights. Qualitative results in Fig.~\ref{fig:in_the_wild_1},~\ref{fig:in_the_wild_2}, and ~\ref{fig:in_the_wild_3} show that diffusion-based methods like Animate Anyone~\cite{AnimateAnyone} and Champ~\cite{Champ} often produce inconsistencies, such as misplaced faces, while IDOL~\cite{idol} tends to generate body shapes that deviate from the reference image and introduces floating artifacts. In contrast, our method preserves identity, maintains high-fidelity details, and generates coherent novel views and poses under real-world conditions, demonstrating strong robustness and versatility. A user study further confirms that our method effectively preserves the reference appearance while aligning with user preferences, with details provided in the following section.

\paragraph{User Study.} 
To validate the perceptual advantages of our method, we conducted a user study with 31 participants, consisting of 20 questions across four aspects: \textbf{Identity \& Appearance Preservation}, \textbf{Novel View Synthesis}, \textbf{Novel Pose Synthesis}, and \textbf{Cross-View Consistency}. Each aspect contained five questions with different subjects, and participants evaluated visual quality, identity preservation, and consistency according to the task in each category. For comparison, IDOL~\cite{idol} was evaluated using its official pre-trained model without additional retraining on our dataset. Its rendered images may have slightly different camera viewpoints, but participants were instructed to focus on visual fidelity and detail preservation, ensuring the user study results remain meaningful. The results of this study are summarized in Tab.~\ref{tab:user_study}. 

We further conducted an in-the-wild user study with 16 participants, consisting of five questions designed to evaluate the robustness of each method under unconstrained scenarios. The study was performed on in-the-wild datasets, including DeepFashion~\cite{Deepfashion_dataset} and SHHQ~\cite{SHHQ_dataset}. Participants compared methods in terms of identity preservation, visual realism, and overall perceptual quality. The results, summarized in Tab.~\ref{tab:user_study_in_the_wild}, demonstrate our method's strong generalizability across diverse, real-world images.

\begin{table*}[]
\centering
\caption{User study results across four aspects: \textbf{Identity \& Appearance Preservation}, \textbf{Novel View Synthesis}, \textbf{Novel Pose Synthesis}, and \textbf{Cross-View Consistency}. Each value indicates the percentage of votes received by each method (higher is better). The best result in each column are highlighted in \textbf{bold}.}
\resizebox{\textwidth}{!}{%
\begin{tabular}{l|cccc}
\toprule
Method & Identity \& Appearance & Novel View & Novel Pose & Cross-View Consistency  \\
\midrule
SHERF & 13.55\% & 0.00\% & 5.16\% & 0.65\% \\
\midrule
Animate Anyone & 5.16\% & 14.19\% & 8.39\% & 16.77\% \\
Champ & 12.90\% & 16.13\% & 8.39\% & 9.03\% \\
\midrule
IDOL & 32.90\% & 7.10\% & 8.39\% & 0.65\% \\
\midrule
\textbf{Ours (Blur2Sharp)} & \textbf{35.48\%} & \textbf{62.58\%} & \textbf{69.68\%} & \textbf{72.90\%} \\
\bottomrule
\end{tabular}%
}
\label{tab:user_study}
\end{table*}


\begin{table}[]
\centering
\caption{In-the-wild user study results. Each value indicates the percentage of votes received by each method (higher is better). The best result is highlighted in \textbf{bold}.}
\resizebox{\columnwidth}{!}{%
\begin{tabular}{c|ccccc}
\toprule
 & SHERF & Animate Anyone & Champ & IDOL & Ours (Blur2Sharp) \\
\midrule
User Preference (\%) & 28.75 & 7.50 & 17.50 & 15.00 & \textbf{31.25} \\
\bottomrule
\end{tabular}%
}
\label{tab:user_study_in_the_wild}
\end{table}

\begin{table}[]
\centering
\caption{Quantitative comparison of novel pose synthesis across 4 views on the HuGe100K dataset~\cite{idol}. The evaluation metrics include PSNR, SSIM, LPIPS, and FID. The best method is highlighted in \textbf{bold}.}
\resizebox{\columnwidth}{!}{%
\begin{tabular}{l|cccc}
\toprule
Method & \multicolumn{4}{c}{HuGe100K}  \\
 & PSNR $\uparrow$ & SSIM $\uparrow$ & LPIPS $\downarrow$ & FID $\downarrow$  \\
\midrule
IDOL & 20.67 & 0.897 & 0.159 & 23.3966  \\
\textbf{Ours (Blur2Sharp)} & \textbf{24.22} & \textbf{0.936} & \textbf{0.042} & \textbf{7.984}  \\
\bottomrule
\end{tabular}%
}
\label{tab:3dgs_comparison}
\end{table}

\end{document}